\documentclass[10pt,twocolumn,letterpaper]{article}

\usepackage{iccv}
\usepackage{times}
\usepackage{epsfig}
\usepackage{graphicx}
\usepackage{amsmath}
\usepackage{amssymb}
\usepackage{amsbsy}
\usepackage{bm}

\DeclareMathOperator*{\argmin}{arg\,min}
\usepackage{booktabs}
\usepackage{subfig}
\usepackage{placeins}
\usepackage{multirow}
\usepackage{varwidth}
\usepackage[skip=-10pt]{caption}
\usepackage{tikz}

\newcommand*\circled[1]{\tikz[baseline=(char.base)]{
 \node[shape=circle,draw,inner sep=0.5pt] (char) {#1};}}

\usepackage[pagebackref=false,breaklinks=true,letterpaper=true,colorlinks,bookmarks=false]{hyperref}

\iccvfinalcopy 

\newcommand{\bj}{\mathbf{j}}
\newcommand{\pose}{\mathbf{p}}
\newcommand{\calD}{\mathcal{D}}

\newcommand{\calT}{\mathcal{T}}

\DeclareMathOperator{\pred}{pred}
\DeclareMathOperator{\synth}{synth}
\DeclareMathOperator{\updater}{updater}
\newcommand{\inp}{\text{input}}

\def\iccvPaperID{999} 

\ificcvfinal\fi
\begin{document}

\title{Training a Feedback Loop for Hand Pose Estimation}

\author{Markus Oberweger \qquad Paul Wohlhart \qquad Vincent Lepetit\\
Institute for Computer Graphics and Vision\\
Graz University of Technology, Austria\\
{\tt\small \{oberweger,wohlhart,lepetit\}@icg.tugraz.at}
}

\maketitle

\begin{abstract}
We propose an entirely data-driven approach to estimating the 3D pose of a hand
given a depth image. We show that we can correct the mistakes made by a
Convolutional Neural Network trained to predict an estimate of the 3D pose by
using a feedback loop. The components of this feedback loop are also Deep
Networks, optimized using training data. They remove the need for fitting a 3D
model to the input data, which requires both a carefully designed fitting function and
algorithm. We show that our approach outperforms 
state-of-the-art methods, and is efficient as our implementation runs at over
400~fps on a single GPU.
\end{abstract}


\section{Introduction}

Accurate hand pose estimation is an important requirement for many Human
Computer Interaction or Augmented Reality tasks~\cite{Erol2007}, and has been
steadily regaining ground as a focus of research interest in the past few
years~\cite{Keskin2011,Keskin2012,Melax2013,Oikonomidis2011,Qian2014,Tang2014,Tang2013,Xu2013},
probably because of the emergence of 3D sensors. Despite 3D sensors, however,
it is still a very challenging problem, because of the vast range of potential
freedoms it involves, and because images of hands exhibit self-similarity and
self-occlusions.

A popular approach is to use a discriminative method to predict the position of the joints~\cite{Keskin2011,Oberweger2015,Tang2014,Tang2013,Tompson2014}, because they are now robust and fast. To refine the pose further, they are often used to initialize an optimization where a 3D model of the hand is fit to the input depth data~\cite{Ballan2012,LaGorce2011,Oikonomidis2011,Plaenkers03,Qian2014,Sharp2015,Sridhar2013,Tzionas2014}. Such an optimization remains complex, however, and typically requires the maintaining of multiple hypotheses~\cite{Oikonomidis2011a,Oikonomidis2011,Qian2014}. It also relies on a criterion to evaluate how well the 3D model fits to the input data, and designing such a criterion is not a simple and straightforward task~\cite{Ballan2012,LaGorce2011,Sridhar2013}.

\begin{figure}
\begin{center}
\includegraphics[width=0.9\linewidth]{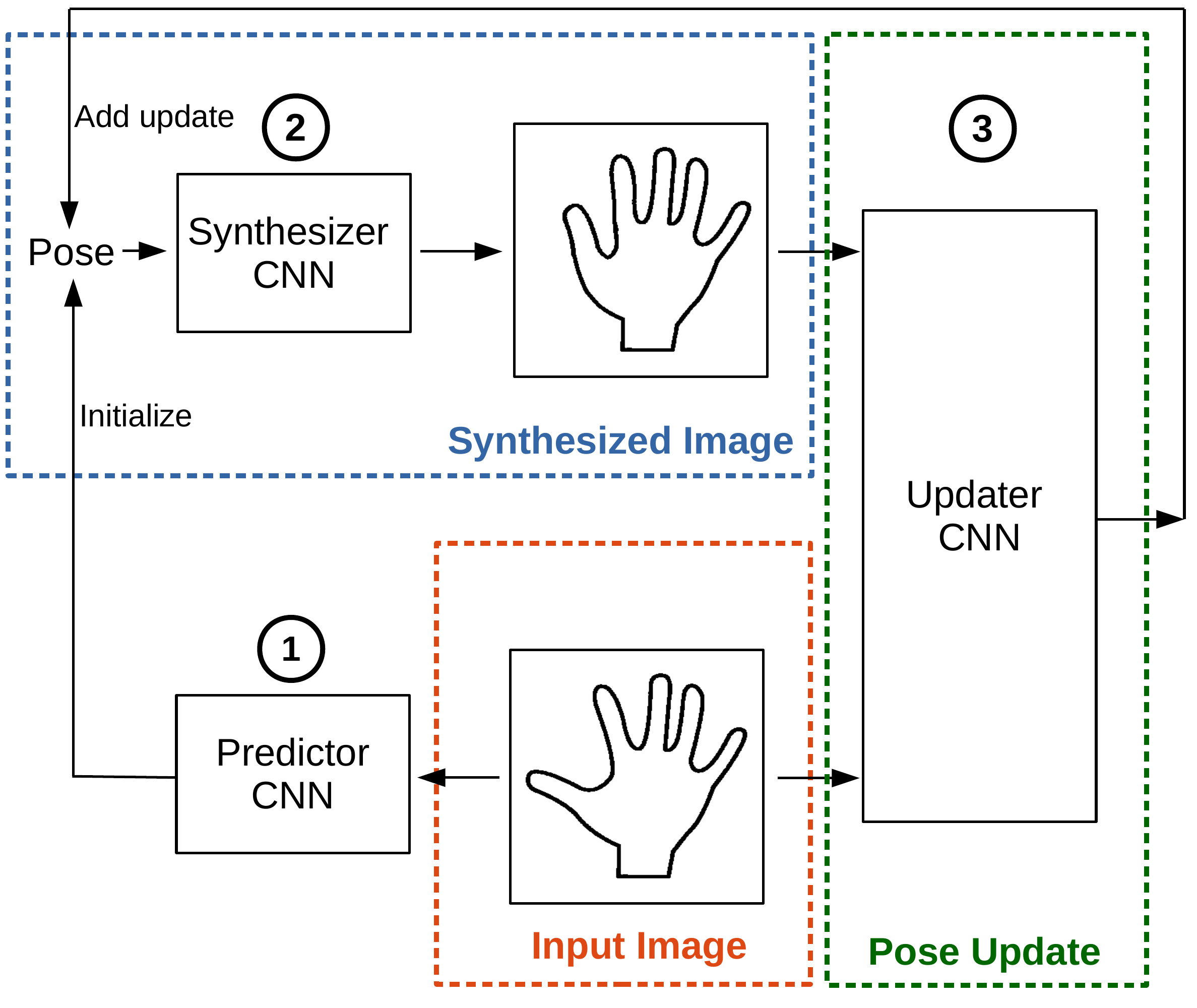}
\end{center}
 \caption{Overview of our method. We use a first CNN \protect\circled{1} to predict an initial
 estimate of the 3D pose given an input depth image of the hand. The pose
 is used to synthesize an image \protect\circled{2}, which is used together with the input depth
 image to derive a pose update \protect\circled{3}. The update is applied to the pose and the
 process is iterated.}
\label{fig:architecture_overview}
\label{fig:teaser}
\end{figure}

In this paper, we first show how we can get rid of the 3D model of the hand altogether and
build instead upon recent work that learns to generate images from
training data~\cite{Dosovitskiy2015}. We then introduce a method that
learns to provide updates for improving the current estimate of the pose, given the
input image and the image generated for this pose estimate as shown in
Fig.~\ref{fig:teaser}. By iterating this method a number of times, we can correct the
mistakes of an initial estimate provided by a simple discriminative method. All
the components are implemented as Deep Networks with simple architectures.

Not only is it interesting to see that all the components needed for hand
registration that used to require careful design can be learned instead, 
but we will also show that our approach has superior performance when compared to the
state-of-the-art methods. It is also efficient; our implementation runs at over
400~fps on a single GPU.

Our approach is related to generative approaches~\cite{Bishop06}, in
particular~\cite{Tang2014a} which also features a feedback loop reminiscent of
ours. However, our approach is deterministic and does not require an
optimization based on distribution sampling, on which generative approaches
generally rely, but which tend to be inefficient.

\section{Related Work}
\label{sec:relatedwork}

Hand pose estimation is a frequently visited problem in Computer Vision, and
is the subject of a plethora of published work. We refer to~\cite{Erol2007} for an overview
of earlier work, and here we will discuss only more recent work, which can be
divided into two types of approach.

The first type of approach is based on discriminative models that
aim at directly predicting the joint locations from RGB or RGB-D
images. Some recent works include \cite{Keskin2011,Keskin2012,Kuznetsova2013,Tang2014,Tang2013} that use different approaches with Random Forests, but restricted to static gestures, showing difficulties with occluded joints, or causing high inaccuracies at finger tips. These problems have been addressed by more recent works of~\cite{Oberweger2015,Tompson2014} that use Convolutional Neural Networks, nevertheless still lacking in high accuracy levels.

Our approach, however, is more related to the second type, which covers
generative, model-based methods. The works of this type are developed from four generic
building blocks: (1) a hand model, (2) a similarity function that measures the
fit of the observed image to the model, (3) an optimization algorithm that
maximizes the similarity function with respect to the model parameters, and
(4) an initial pose from which the optimization starts.

For the hand model, different hand-crafted models were proposed. The choice of a
simple model is important for maintaining real-time capabilities, and representing a
trade-off between speed and accuracy as a response to a potentially high degree of model
abstraction. Different hand-crafted geometrical approximations for hand models
were proposed.
For example, \cite{Qian2014} uses a hand model consisting of spheres, 
\cite{Oikonomidis2011} adds cylinders, ellipsoids, and cones.
\cite{Sridhar2013} models the hand
from a Sum of Gaussians. More holistic hand representations are used
by~\cite{Ballan2012,Sharp2015,Tzionas2014}, with a linear blend skinning model
of the hand that is rendered as depth map. \cite{Xu2013} increases the matching
quality by using depth images that resemble the same noise pattern as the depth
sensor. \cite{LaGorce2011} uses a fully shaded and textured triangle mesh controlled by a skeleton.

Different modalities were proposed for the similarity function, which are
coupled to the used model.
The modalities include, \eg, depth values \cite{Melax2013,Oikonomidis2011a,Qian2014}, 
salient points, edges, color \cite{LaGorce2011}, or 
combinations of these \cite{Ballan2012,Oikonomidis2011,Sridhar2013,Tzionas2014}.

The optimization of the similarity function is a critical part, as the high
dimensional pose space is prone to local minima. Thus, Particle Swarm
Optimization is often used to handle the high dimensionality of the pose
vector~\cite{Oikonomidis2011a,Oikonomidis2011,Qian2014,Sharp2015}. Differently,
\cite{Ballan2012,LaGorce2011,Sridhar2013} use gradient-based methods
to optimize the pose, while \cite{Melax2013} uses dynamics simulation.
Due to the high computation time of these optimization methods, which have to be
solved for every frame, \cite{Xu2013} does not optimize the pose but only
evaluates the similarity function for several proposals to select the best fit.

In order to kick-start optimization, \cite{Sridhar2013} uses a
discriminative part-based pose initialization, and \cite{Qian2014}
uses finger tips only. \cite{Xu2013} predicts candidate poses using a Hough
Forest. \cite{LaGorce2011} requires predefined hand color and position, and
\cite{Oikonomidis2011} relies on a manual initialization. Furthermore,
tracking-based methods use the pose of the last 
frame~\cite{Melax2013,Oikonomidis2011,Tzionas2014}, which can be
problematic if the difference between frames is too large, because of fast
motion or low frame rates.

Our approach differs from previous work in the first three building blocks.
We do not use a deformable CAD model of the hand. Instead, we learn from
registered depth images to generate realistic depth images of hands, similar to
work on inverse graphics networks~\cite{Dosovitskiy2015,Kulkarni2015}, and other
recent work on generating
images~\cite{Goodfellow2014,Kulkarni2014}. This approach is very
convenient, since deforming correctly and rendering a CAD model of the hand in a realistic manner requires a significant input of engineering work.

In addition, we do not use a hand-crafted similarity function and an optimization
algorithm. We learn rather to predict updates that improve
the current estimate of the hand pose from training data, given the input depth image and a
generated image for this estimate. Again this approach is very convenient, since it means 
we do not need to design the similarity function and the optimization algorithm,
neither of which is a simple task.

Since we learn to generate images of the hand, our approach is also
related to generative approaches, in particular~\cite{Tang2014a}.
It uses a feedback loop with an updater mechanism akin to ours.
It predicts updates for the position from which a patch is
cropped from an image, such that the patch fits best to the
output of a generative model. However, this step does not
predict the full set of parameters. The hidden states of the model
are found by a costly sampling process.

\cite{Nair2008} relies on a given black-box image synthesizer to provide
synthetic samples on which the regression network can be trained. It then
learns a network to substitute the black-box graphics model, which can
ultimately be used to update the pose parameters to generate an image that most
resembles the input. In contrast, we learn the generator model directly from
training data, without the need for a black-box image synthesizer. Moreover, we
will show that the optimization is prone to output infeasible poses or get stuck
in local minima and therefore introduce a better approach to improve the pose.

\section{Model-based Pose Optimization}
\label{sec:main}

In this section, we will first give an overview of our method. We
then describe in detail the different components of our method: A
discriminative approach to predict a first pose estimate, a method
able to generate realistic depth images of the hand, and a learning-based
method to refine the initial pose estimate using the generated depth images.

\subsection{Method Overview}
\label{sec:overview}

Our objective is to estimate the pose $\pose$ of a hand in the form
of the 3D locations of its joints $\pose = \{\bj_i\}^J_{i=1}$ with
$\bj_i=(x_i,y_i,z_i)$ from a single depth image
$\calD$. In practice, $J=14$ for the dataset we use. We assume that a
training set $\calT = \{(\calD_i, \pose_i)\}^N_{i=1}$ of depth images labeled
with the corresponding 3D joint locations is available.

As explained in the introduction, we first train a predictor to predict an initial pose estimate
$\widehat\pose^{(0)}$ in a discriminative manner given an input depth image $\calD_\inp$:
\begin{equation}
\widehat\pose^{(0)} = \pred(\calD_\inp) \;\; .
\end{equation}
We use a Convolutional Neural Network (CNN) to implement the $\pred(.)$ function
with a standard architecture. More details will be given in
Section~\ref{sec:pred}.

\begin{figure}
\begin{center}
\includegraphics[width=0.24\linewidth]{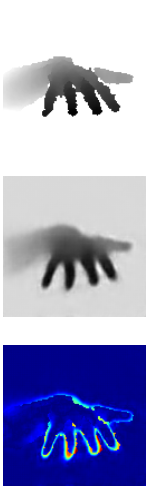}
\includegraphics[width=0.24\linewidth]{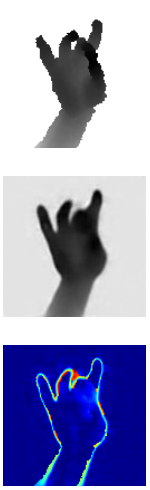}
\includegraphics[width=0.24\linewidth]{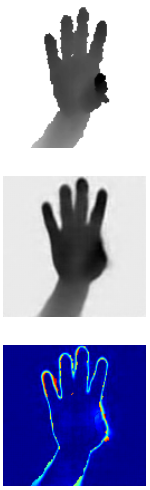}
\includegraphics[width=0.24\linewidth]{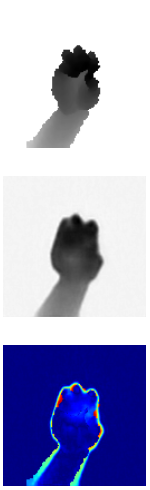}
\end{center}
 \caption{Samples generated by the synthesizer for different poses from the
 test set. \textbf{Top:} Ground truth depth image. \textbf{Middle:}
 Synthesized depth image using our learned hand model. \textbf{Bottom:}
 Color-coded, pixel-wise difference between the depth images. The synthesizer
 is able to render convincing depth images for a very large range of poses.
 The largest errors are located near the occluding contours of the hand,
 which are noisy in the ground truth images. (Best viewed on screen)}
\label{fig:gensamples}
\end{figure}

In practice, $\widehat\pose^{(0)}$ is never perfect, and following the motivation provided in the
introduction, we introduce a hand model learned from the training data. As
shown in Fig.~\ref{fig:gensamples}, this model can synthesize the depth image
corresponding to a given pose $\pose$, and we will refer to this model as the
\emph{synthesizer}:
\begin{equation}
\calD_{\synth} = \synth(\pose) \;\; .
\end{equation}
We also use a Deep Network to implement the synthesizer.

\begin{figure}
\begin{center}
\subfloat[]{\includegraphics[width=0.24\linewidth]{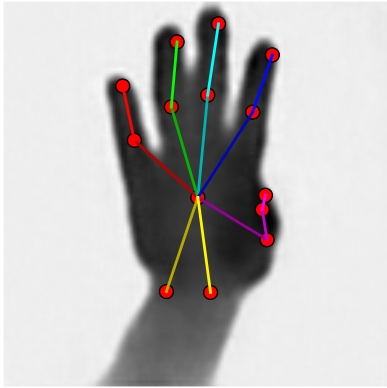}}
\subfloat[]{\includegraphics[width=0.24\linewidth]{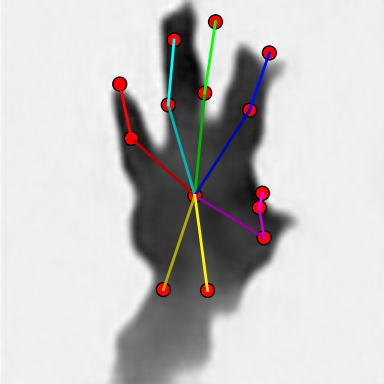}}
\subfloat[]{\includegraphics[width=0.24\linewidth]{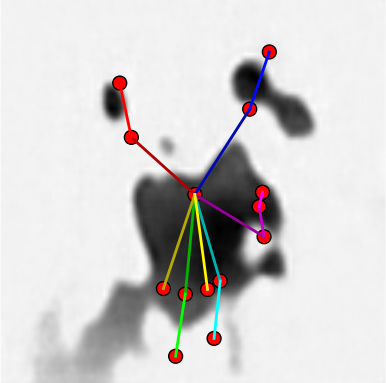}}
\subfloat[]{\includegraphics[width=0.24\linewidth]{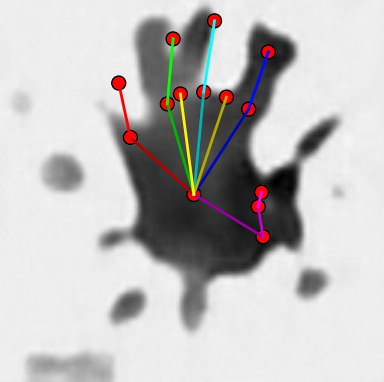}}
\end{center}
 \caption{Synthesized images for physically impossible poses. Note
 the colors that indicate different fingers. (a) shows a feasible
 pose with its synthesized image. (b) shows the synthesized image
 for the same pose after swapping the ring and middle finger positions. In (c) the
 ring and middle finger are flipped downwards, and in (d) the
 wrist joints are flipped upwards. (Best viewed on screen)}
\label{fig:gensamples_impossible}
\end{figure}

A straightforward way of using this synthesizer would consist in estimating the hand pose
$\widehat\pose$ by minimizing the squared loss between the input image and the
synthetic one: 
\begin{equation}
\widehat\pose = \argmin_\pose \lVert \calD_\inp - \synth(\pose) \rVert^2 \;\; .
\label{eq:naive}
\end{equation}
This is a non-linear least-squares problem, which can be solved iteratively
using $\widehat\pose^{(0)}$ as initial estimate. However, the objective
function of Eq.~\eqref{eq:naive} exhibits many local minima. Moreover, during
the optimization of Eq.~\eqref{eq:naive}, $\pose$ can take values that
correspond to physically infeasible poses. For such values, the output
$\synth(\pose)$ of the synthesizer is unpredictable as depicted in
Fig.~\ref{fig:gensamples_impossible}, and this is likely to make the optimization of
Eq.~\eqref{eq:naive} diverge or be stuck in a local minimum, as we will show
in the experiments in Section~\ref{sec:naive}.

We therefore introduce a third function that we call the $\updater(.,.)$. It learns to predict updates, which are applied to the pose estimate to improve it, given the input image $\calD_\inp$ and the image $\synth(\pose)$ produced by the synthesizer:
\begin{equation}
\widehat\pose^{(i+1)} \;\leftarrow\; \widehat\pose^{(i)} + \updater(\calD_\inp, \synth(\widehat\pose^{(i)})) \;\; .
\label{eq:updater}
\end{equation}
We iterate this update several times to improve the initial pose
estimate. Again, the $\updater(.,.)$ function is implemented as a Deep Network.

We detail below how we implement and train the $\pred(.)$, $\synth(.)$ and $\updater(.,.)$
functions.

\subsection{Learning the Predictor Function $\pred(.)$}
\label{sec:pred}

The predictor is implemented as a CNN. The network consists of a convolutional layer
with $5\times 5$ filter kernels producing 8 feature maps. These feature maps are
max-pooled with $4\times 4$ windows, followed by a
hidden layer with 1024 neurons and an output layer with one neuron for each
joint and dimension, \ie ${3\cdot J}$ neurons. The predictor is parametrized by
$\Phi$, which is obtained by minimizing
\begin{equation}
\label{eq:train_pred}
\widehat\Phi = \argmin_{\Phi} \sum_{(\calD, \pose) \in \calT} \lVert \pred_\Phi(\calD) - \pose \rVert^2_2 + \gamma\lVert\Phi\rVert^2_2\;\; ,
\end{equation}
where the second term is a regularizer for weight decay with $\gamma=0.001$.

\subsection{Learning the Synthesizer Function $\synth(.)$}

\begin{figure*}[t]
\begin{center}
\includegraphics[width=0.9\linewidth]{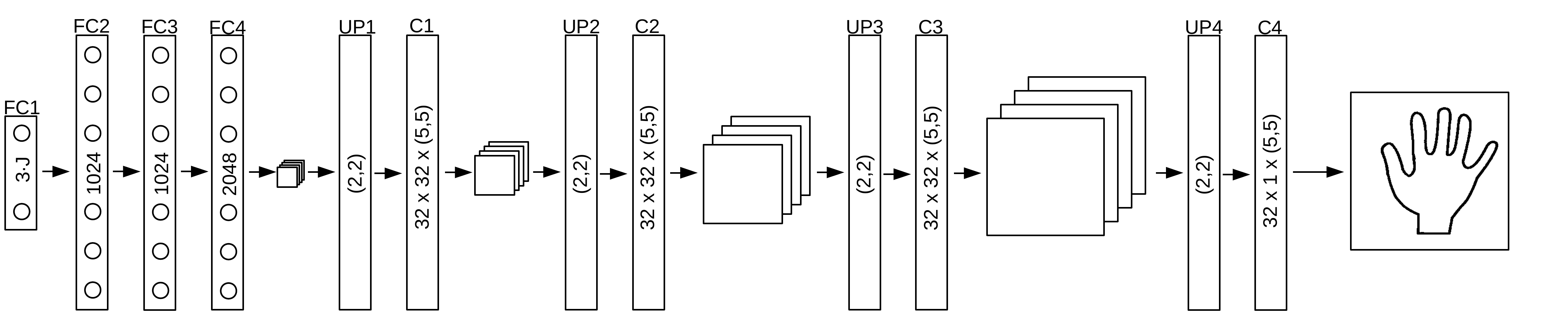}
\end{center}
 \caption{Network architecture of the synthesizer used to generate depth
 images of hands given their poses. The input of the network is the hand
 pose. The fully connected hidden layers create a 2048 dimensional latent representation 
 at \textsf{FC4} which is reshaped into 32 feature maps of size $8\times 8$.
 The feature maps are gradually enlarged by successive
 unpooling and convolution operations. The last convolution layer combines
 the feature maps to derive a single depth image of size $128 \times 128$. All layers have rectified-linear units, except the last layer which has linear units. \textsf{C} denotes a
 convolutional layer with the number of filters and the filter size
 inscribed, \textsf{FC} a fully connected layer with the number of neurons,
 and \textsf{UP} an unpooling layer with the upscaling factor.}
\label{fig:architecture}
\end{figure*}

We use a CNN to implement the synthesizer, and we train it using the set $\calT$
of annotated training pairs. The network architecture is strongly inspired
by~\cite{Dosovitskiy2015}, and is shown in Fig.~\ref{fig:architecture}. It
consists of four hidden layers, which learn an initial latent representation of
feature maps apparent after the fourth fully connected layer \textsf{FC4}.
These latent feature maps are followed by several unpooling and convolution
layers. The unpooling operation, used for example
by~\cite{Dosovitskiy2015,Goodfellow2014,Zeiler2014,Zeiler2011}, is the inverse
of the max-pooling operation: The feature map is expanded, in our case by a
factor of 2 along each image dimension. The emerging "holes" are filled with
zeros. The expanded feature maps are then convolved with trained 3D filters to
generate another set of feature maps. These unpooling and convolution operations
are applied subsequently. The last convolution layer combines all feature maps
to generate a depth image.

We learn the parameters $\widehat\Theta$ of the network by minimizing
the difference between the generated depth images
$\synth(\pose)$ and the training depth images $\calD$ as
\begin{equation}
\label{eq:train_synth}
\widehat\Theta = \argmin_{\Theta} \sum_{(\calD, \pose) \in \calT} \frac{1}{|\calD|} \lVert \synth_\Theta(\pose) - \calD \rVert^2_2 \;\; .
\end{equation}
We perform the optimization in a layer-wise fashion. We start by training
the first $8\times 8$ feature map. Then we gradually extend the
output size by adding another unpooling and convolutional layer and train
again, which achieves lesser errors than end-to-end training in our
experience.

The synthesizer is able to generate accurate images, maybe surprisingly well for
such a simple architecture. The median pixel error on the test set is only
1.3~mm. However, the average pixel error is $10.9\pm32.2mm$. This is mostly due
to noise in the input images along the outline of the hand, which is smoothed
away by the synthesizer. The average depth accuracy of the sensor is $\pm
1~mm$~\cite{Primesense2015}.

\subsection{Learning the Updater Function $\updater(.,.)$}

\begin{figure*}[tb]
\begin{center}
\includegraphics[width=0.7\linewidth]{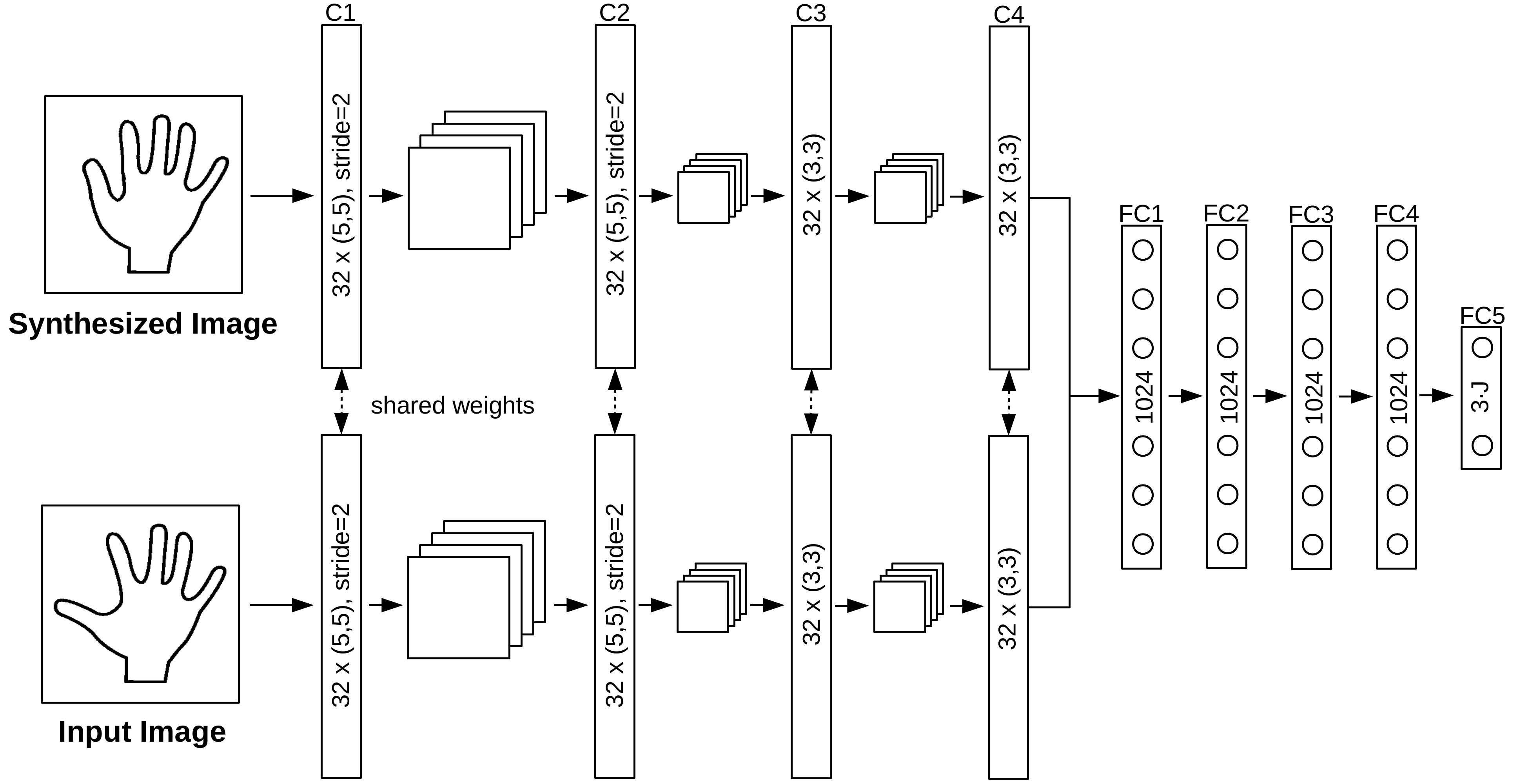}
\end{center}
 \caption{Architecture of the updater. The network consists of two identical
 paths with shared weights that contain several convolutional layers that
 use a filter stride to reduce the size of the feature maps. The final
 feature maps are concatenated and fed into a fully connected network. All
 layers have rectified-linear units, except the last layer which has linear
 units. The pose update is used to refine the initial pose and the refined
 pose is again fed into the synthesizer to iterate the whole procedure. As
 in Fig.~\ref{fig:architecture}, \textsf{C} denotes a convolutional layer,
 and \textsf{FC} a fully connected layer.}
\label{fig:architecture_refine_comp}
\label{fig:updater}
\end{figure*}

The updater function $\updater(.,.)$ takes two depth images as input. As
already stated in Eq.~\eqref{eq:updater}, at run-time, the first image is the
input depth image, the second image is the image returned by the synthesizer for
the current pose estimate. Its output is an update that improves the pose
estimate. The architecture, shown in Fig.~\ref{fig:updater}, is 
 inspired by the Siamese network~\cite{Chopra2005}. It consists of two
identical paths with shared weights. One path is fed with the observed image
and the second path is fed with the image from the synthesizer. Each path
consists of four convolutional layers. We do not use max-pooling here, but a
filter stride~\cite{Jain2014,Liu2015} to reduce the resolution of the feature maps. We experienced inferior
accuracy with max-pooling compared to that with stride, probably because max-pooling
introduces spatial invariance~\cite{Scherer2010} that is not desired for this
task. The feature maps of the two paths are concatenated and fed into a fully
connected network that outputs the update.

\begin{figure}
\begin{center}
\includegraphics[width=0.6\linewidth]{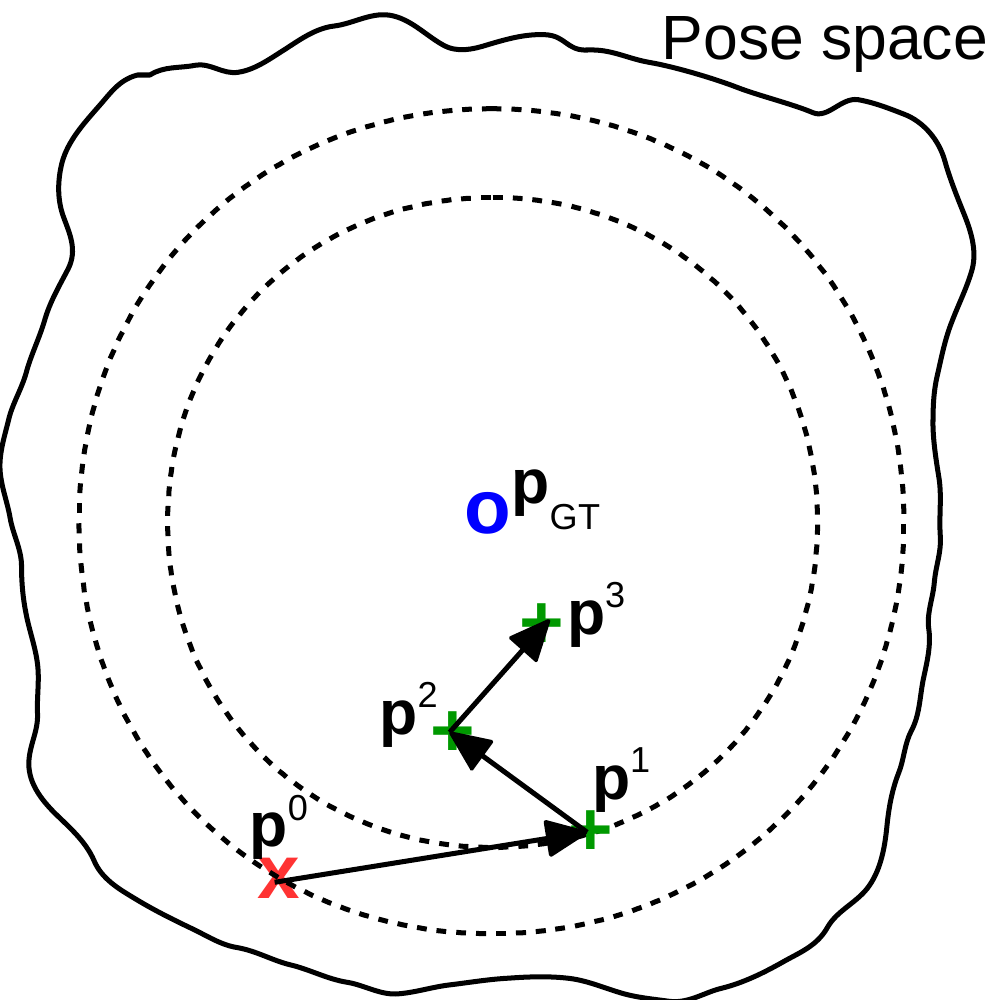}
\end{center}
 \caption{Our iterative pose optimization in high-dimensional space,
 schematized here in 2D. We start at an initial pose
 (\textcolor{red}{$\bm{\times}$}) and want to converge to the ground truth
 pose (\textcolor{blue}{$\bm{\circ}$}), that maximizes image similarity.
 Our predictor generates updates for each pose (\textcolor{green}{$\bm{+}$})
 that bring us closer. The updates are predicted from the synthesized image of the current pose
 estimate and the observed depth image. (Best viewed in color)}
\label{fig:iterative}
\end{figure}

Ideally, the output of the updater should bring the pose estimate to the correct pose in a
single step. However, this is a very difficult problem, and we could not get the network to reduce the initial training error within a reasonable timeframe. 
However, our only requirement from the updater is for it to
output an update which brings us closer to the ground truth as shown in Fig.~\ref{fig:iterative}. We iterate
this update procedure to get closer step-by-step. Thus, the update should
follow the inequality
\begin{equation}\label{eq:train_refine2}
	\lVert \pose + \updater(\calD, \synth(\pose)) -
 \pose_\text{GT} \rVert_2 < \lambda \lVert \pose
 -\pose_\text{GT} \rVert_2 \;\; ,
\end{equation}
where $\pose_\text{GT}$ is the ground truth pose for image $\calD$, and ${\lambda
\in [0,1]}$ is a multiplicative factor that specifies a minimal improvement. We
use $\lambda = 0.6$ in our experiments. 

We optimize the parameters $\Omega$ of the updater by minimizing the following cost function
\begin{equation}\label{eq:train_refine}
	\mathcal{L} = \sum_{(\calD,\pose)\in\calT} \sum_{\pose'\in \calT_{\calD}} \max(0,\lVert \pose'' - \pose \rVert_2
 - \lambda \lVert \pose' -\pose \rVert_2) \;\; , 
\end{equation}
where $\pose'' = \pose' + \updater_\Omega(\calD,\synth(\pose'))$, and
$\calT_\calD$ is a set of poses. The introduction of the synthesizer allows us
to virtually augment the training data and add arbitrary poses to
$\calT_{\calD}$, which the updater is then trained to correct.

The set $\calT_{\calD}$ contains the ground truth $\pose$, for which the updater
should output a zero update. We further add as many meaningful deviations from
that ground truth as possible, which the updater might perceive during testing and be
asked to correct. We start by including the output pose of the predictor
$\pred(\calD)$, which during testing is used as initialization of the update
loop. Additionally, we add copies with small Gaussian
noise for all poses. This creates convergence basins around the ground truth, in which the
predicted updates point towards the ground truth, as we show in the evaluation,
and further helps to explore the pose space.

After every 2 epochs, we further augment the set by applying the current updater to
the poses in $\calT_\calD$, that is, we add the set
\begin{equation}
\{\pose_2 \;|\; \exists \pose \in \calT_\calD \text{ s.t. } \pose_2 = \pose + \updater(\calD, \synth(\pose)) \} \; 
\end{equation}
to $\calT_\calD$. This forces the updater to learn to further improve on its own outputs.

In addition, we sample from the current distribution of errors across all the samples,
and add these errors to the poses, thus explicitly focusing the training on common deviations.
This is different from the Gaussian noise and helps to predict correct
updates for larger initialization errors.


\section{Evaluation}
\label{sec:eval}

In this section we evaluate our proposed method on the NYU Hand Pose
Dataset~\cite{Tompson2014}, a challenging real-world benchmark for hand pose
estimation. First, we describe how we train the networks. Then, we introduce
the evaluation metric and the benchmark dataset. Furthermore we evaluate our method qualitatively and quantitatively.

\subsection{Training}
We optimize the networks' parameters by using error back-propagation and apply
the rmsprop~\cite{Tieleman2012} algorithm. We choose a decay parameter of $0.9$
and truncate the gradients to $0.01$. The batch size is $64$. The learning
rate decays over the epochs and starts with $0.01$ for the predictor and
synthesizer, and with $0.001$ for the updater. The networks are trained for
$100$ epochs.

\subsection{Hand Detection}
We extract a fixed-size metric cube from the depth image around the
hand. The depth values within the cube are resized to a $128\times128$
patch and normalized to $[-1,1]$. The depth values are clipped to the
cube sides front and rear. Points for which the depth is
undefined---which may happen with structured light sensors for
example---are assigned to the rear side. This preprocessing step
was also done in~\cite{Tang2014} and provides invariance to different hand-to-camera distances.

\subsection{Benchmark}
We evaluated our method on the NYU Hand Pose Dataset~\cite{Tompson2014}. This
dataset is publicly available, it is backed up by a huge quantity of annotated samples together with very accurate annotations. It also shows a high variability of poses, however, which can
make pose estimation challenging. While the ground truth contains $J=36$
annotated joints, we follow the evaluation protocol
of~\cite{Oberweger2015,Tompson2014} and use the same subset of $J=14$ joints.

The training set contains samples of one person, while the test set has samples
from two persons. The dataset was captured using a structured light RGB-D
sensor, the Primesense Carmine 1.09, and contains over 72k training and 8k test
frames. We used only the depth images for our experiments. They exhibit typical artifacts of structured light sensors: The outlines are noisy and there are missing depth
values along occluding boundaries.

\begin{figure}[t]
\begin{center}
\includegraphics[width=0.49\linewidth]{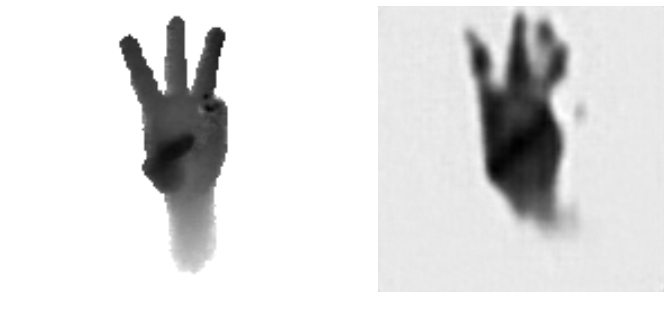}
\includegraphics[width=0.49\linewidth]{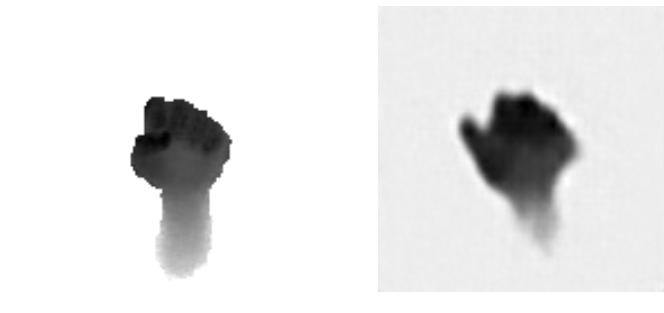}
\end{center}
 \caption{The effects of erroneous annotations in the training data of~\cite{Tang2014}. The figures show input images together with synthesized images which exhibit very blurry outlines and ``ghost'' fingers. The synthesizer learns the erroneous annotations and interpolates between inconsistent ones, causing such artifacts that limit the applicability in our method. (Best viewed on screen)}
\label{fig:results_erroneous_annotation}
\end{figure}

We also considered other datasets for this task; unfortunately, no
further dataset was suitable. The dataset of Tang~\etal~\cite{Tang2014} has
large annotation errors, that cause blurry outlines and ``ghost'' fingers in the synthesized images as shown in Fig.~\ref{fig:results_erroneous_annotation}, which are not
suitable for our method. The datasets of~\cite{Qian2014,Sridhar2013,Xu2013}
provide too little training data to train meaningful models.

\subsection{Comparison with Baseline}

\begin{figure*}[t]
\begin{center}
\subfloat[]{
 \includegraphics[width=0.49\linewidth]{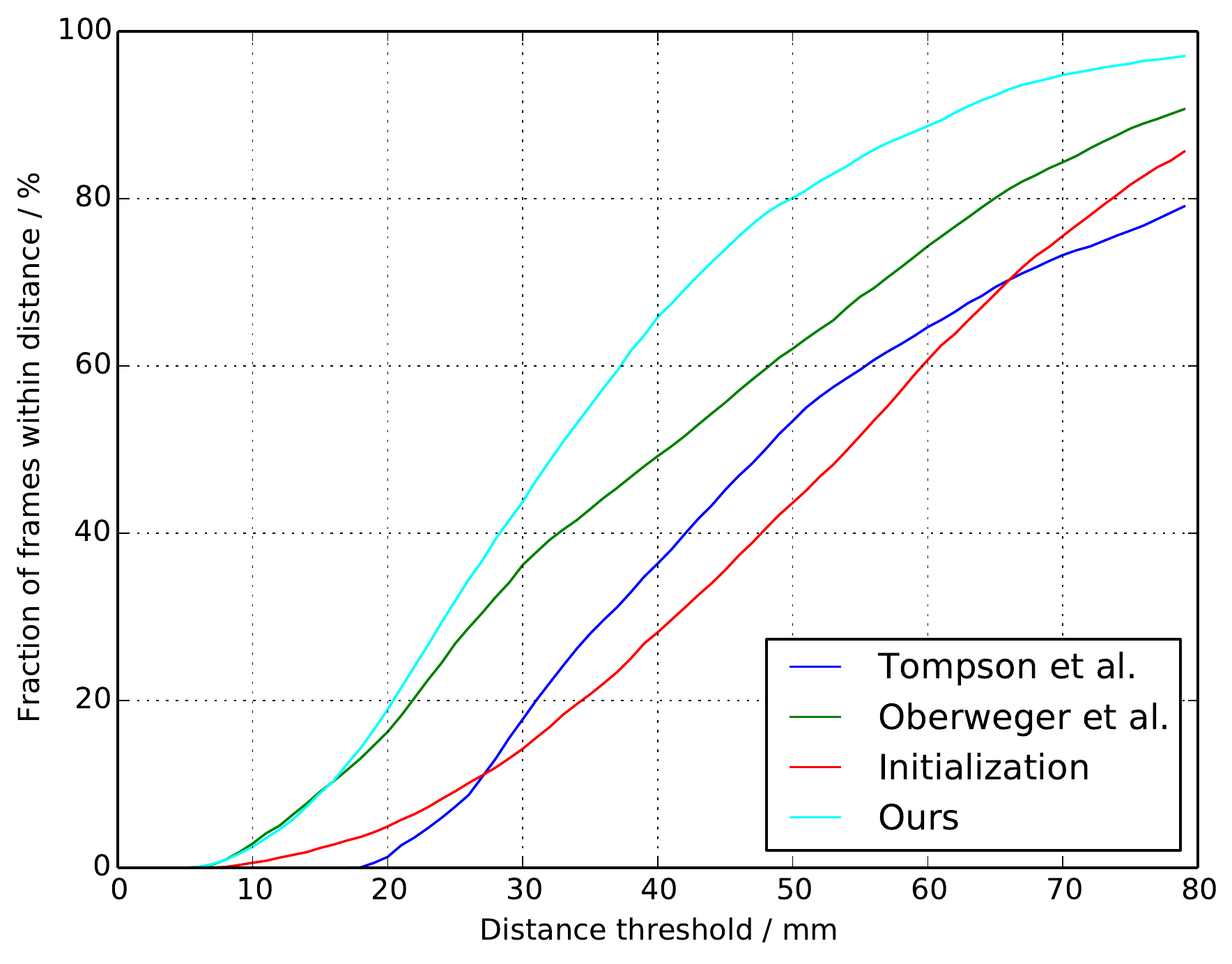}\label{fig:eval_frameswithin_REF}
}
\subfloat[]{
 \includegraphics[width=0.49\linewidth]{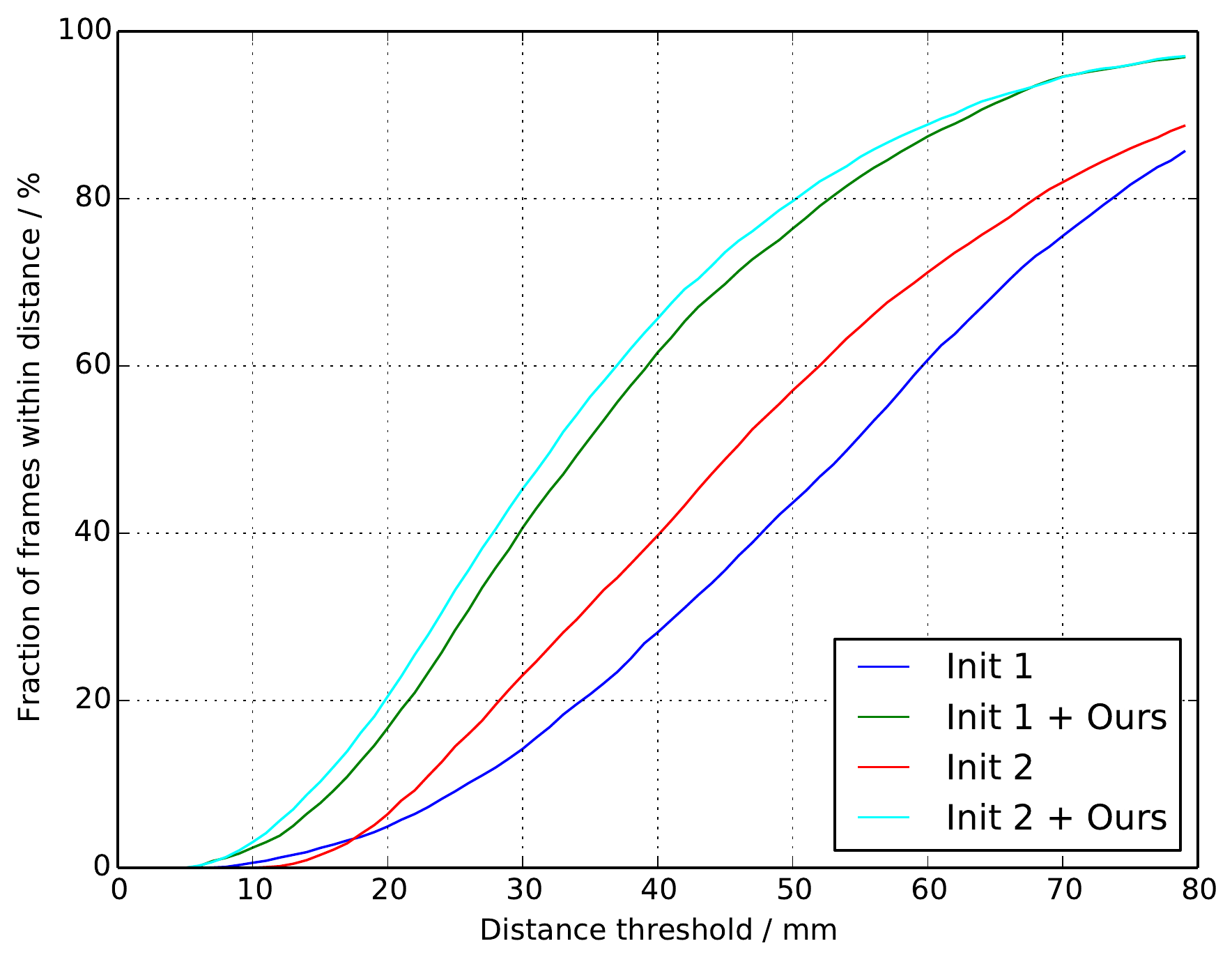}\label{fig:eval_frameswithin_INIT}
}
\end{center}
 \caption{Quantitative evaluation of pose optimization. Both figures show the
 fraction of frames where all joints are within a maximum distance. A
 higher area under the curve denotes better results. In (a) we compare our method to the baseline of Tompson~\etal~\cite{Tompson2014} and Oberweger~\etal~\cite{Oberweger2015}. Although our initialization is worse than both baselines, we can boost the accuracy of the joint locations using our proposed method. In (b) we compare different initializations. \textit{Init 1} is the simple predictor presented in this work. We also use the more sophisticated model of~\cite{Oberweger2015}, denoted as \textit{Init 2}, for a more accurate initialization. The better initialization helps obtain slightly more accurate results, however, our much simpler and faster predictor is already sufficient for our method as initialization. (Best viewed on screen)}
\label{fig:results_quantitative}
\end{figure*}

We show the benefit of using our proposed feedback loop to increase the accuracy of the 3D joint localization.
For this, we compare to Tompson~\etal~\cite{Tompson2014}, and to
Oberweger~\etal~\cite{Oberweger2015}. For \cite{Tompson2014}, we augment their
2D joint locations with the depth from the depth images, as done
in~\cite{Tompson2014}. In case this estimate is outside the hand cube we assign ground truth depth, thus favorably mitigating large errors in those cases.
For \cite{Oberweger2015}, we use
their best CNN that incorporates a 30D pose embedding.

The quantitative comparison is shown in Fig.~\ref{fig:eval_frameswithin_REF}. 
It shows results using the metric of~\cite{Taylor2012} which is generally
regarded as being very challenging. It denotes the fraction of test samples that have all
predicted joints below a given maximum Euclidean distance from the ground
truth. Thus a single erroneous joint results in the deterioration of the whole hand pose.

While the baseline of~\cite{Tompson2014} and~\cite{Oberweger2015} have an average Euclidean joint error of 21~mm and 20~mm respectively, our proposed method reaches an error reduction to 16.5~mm, thus achieving state-of-the-art on the dataset. The initialization with the simple and efficient proposed predictor has an error of 27~mm. We further show an evaluation of different initializations in Fig.~\ref{fig:eval_frameswithin_INIT}. When we use a more complex initialization~\cite{Oberweger2015} with an error of 23~mm, we can decrease the average error to 16~mm. For a more detailed breakdown of the error evaluation of each joint, we refer to the supplementary material.

\subsection{Image-Based Hand Pose Optimization}
\label{sec:naive}

We mentioned in Section~\ref{sec:overview} that the attempt may be made
to estimate the pose by directly optimizing the squared loss between
the input image and the synthetic one as given in Eq.~\eqref{eq:naive} and we
argued that this does not in fact work well. We now demonstrate this empirically.

We used the powerful L-BFGS-B algorithm~\cite{Byrd1995}, which is a box
constrained optimizer, to solve Eq.~\eqref{eq:naive}. We set the constraints on
the pose in such a manner that each joint coordinate stays inside the hand cube.

\begin{figure}
\begin{center}
 \includegraphics[width=0.95\linewidth]{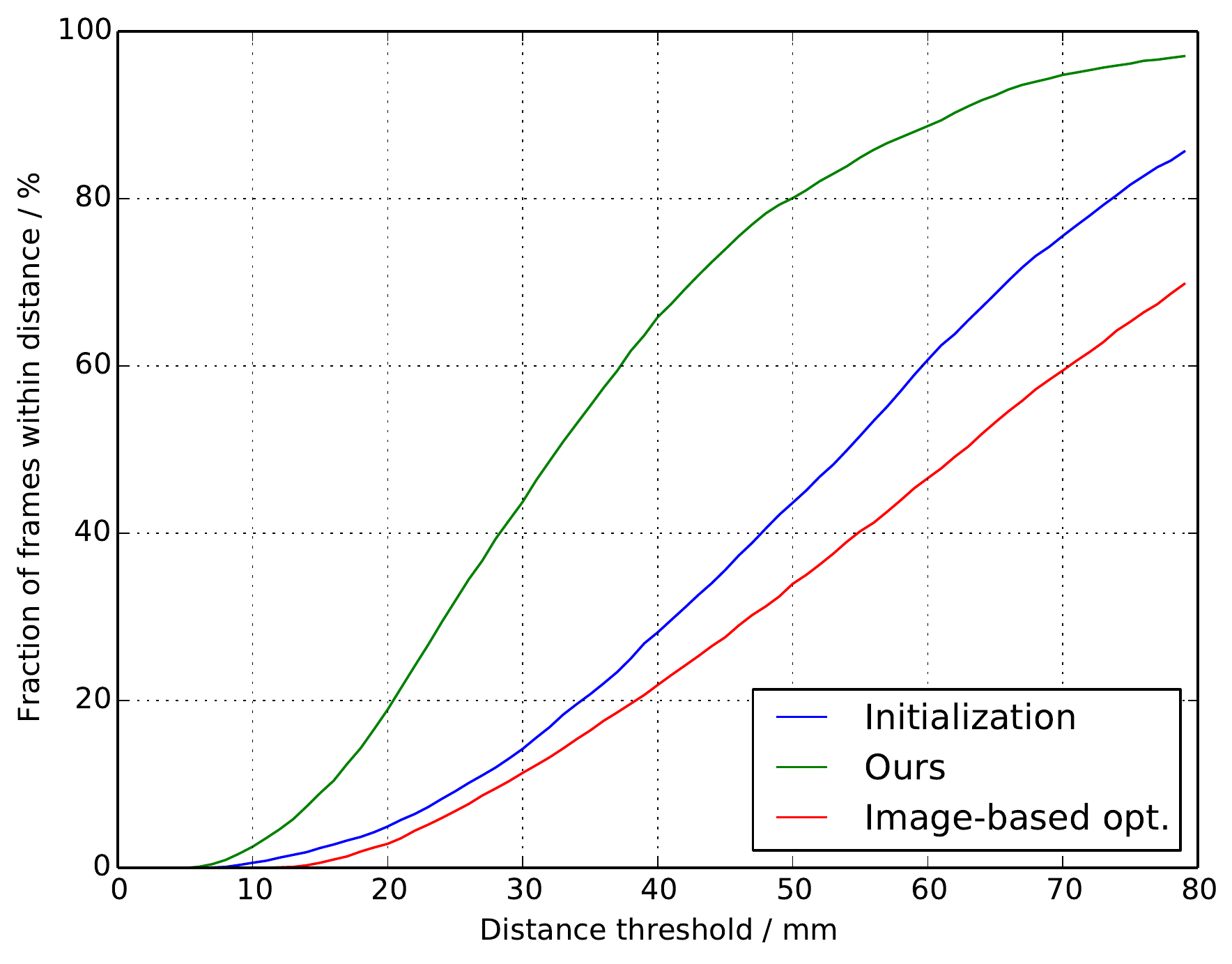}\label{fig:eval_frameswithin_NYU}
\end{center}
 \caption{Comparing image-based optimization and our feedback loop. A higher
 area under the curve denotes better results. The image-based optimization
 actually tends to result in a deterioration of the initial pose estimate. Our proposed
 optimization method performs significantly better. (Best viewed in
 color)}
\label{fig:results_quantitative_pose}
\end{figure}

The minimizer of Eq.~\eqref{eq:naive}, however, does not correspond to a better
pose in general, as shown in Fig.~\ref{fig:gensamplesBFGS}. Although the generated image
looks very similar to the input image, the pose does not improve, moreover it even often becomes worse. Several reasons can account for this. The depth input image
typically exhibits noise along the contours, as in the example 
of Fig.~\ref{fig:gensamplesBFGS}. After several iterations of L-BFGS-B, the
optimization may start corrupting the pose estimate with the result that the synthesizer generates artifacts that fit the noise. As shown in Fig.~\ref{fig:results_quantitative_pose},
the pose after optimization is actually worse than the initial pose
of the predictor.

Furthermore the optimization is prone to local minima due to a noisy error
surface~\cite{Qian2014}. However, we also tried Particle Swarm
Optimization~\cite{Oikonomidis2011a,Qian2014,Sharp2015} a genetic algorithm
popular for hand pose optimization, and obtained similar results. This tends to
confirm that the bad performance comes from the objective function of
Eq.~\eqref{eq:naive} rather than the optimization algorithm.

\begin{figure}
\begin{center}
 \includegraphics[width=0.75\linewidth]{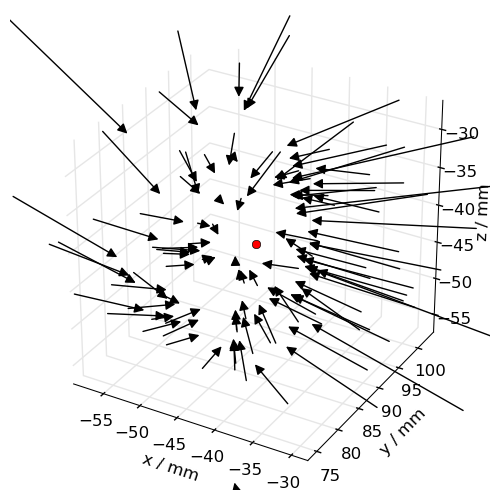} 
\end{center}
 \caption{Predicted updates around ground truth joint position, denoted as \protect\tikz\protect\draw[red,fill=red] (0,0) circle (.5ex);. We initialize noisy joint locations around the ground truth location and show the pose updates as vectors predicted by our updater. The start and end of the vectors denote the initial and the updated 3D joint location. The different updates bring us closer to the ground truth joint location. (Best viewed
 on screen)}
\label{fig:results_quiver}
\end{figure}

By contrast, in Fig.~\ref{fig:results_quiver} we show the predicted updates for different initializations around the ground truth joint location with our updater. It predicts updates that move the pose closer to the ground truth, for almost all initializations.

\bgroup
\setlength{\tabcolsep}{1pt}
\begin{figure}[t]
\begin{center}
\begin{tabular}{@{}ccccc@{}}
Init & Init & Iter 1 & Iter 2 & Final \\
\multirow{2}{*}[2em]{\includegraphics[width=0.19\linewidth]{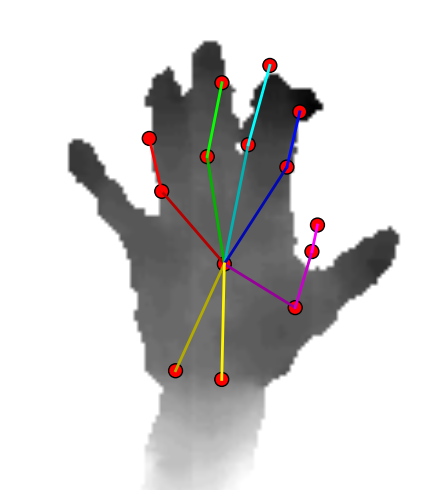}} &
\includegraphics[width=0.19\linewidth]{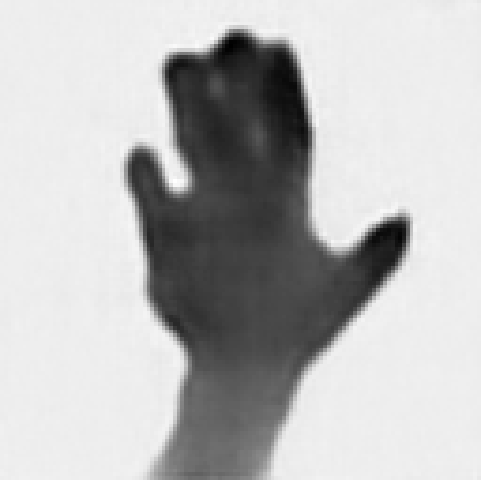} &
\includegraphics[width=0.19\linewidth]{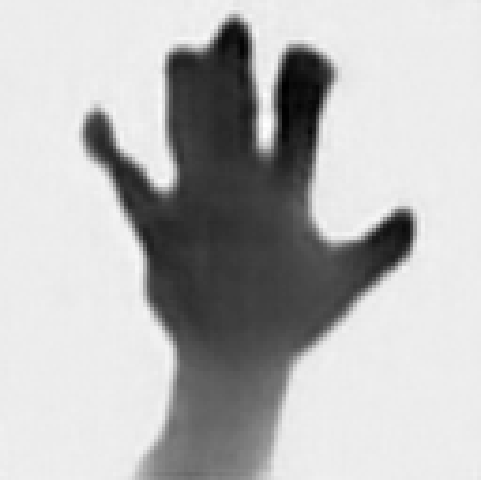} &
\includegraphics[width=0.19\linewidth]{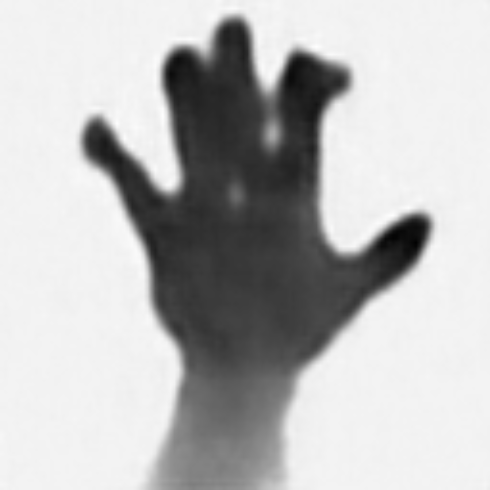} &
\includegraphics[width=0.18\linewidth]{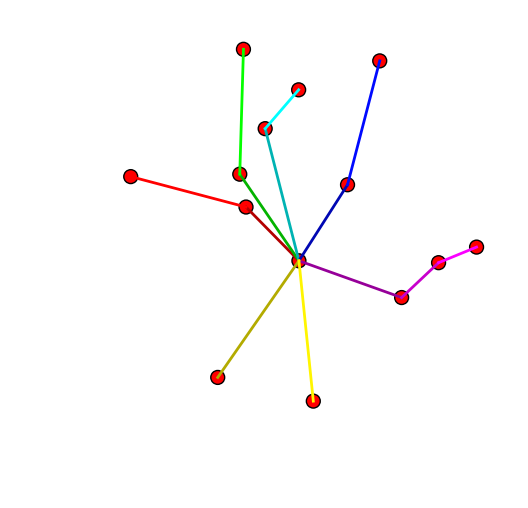} \\
 &
\includegraphics[width=0.19\linewidth]{NYU_GENREF_GENREF_2451_ITER_0} &
\includegraphics[width=0.19\linewidth]{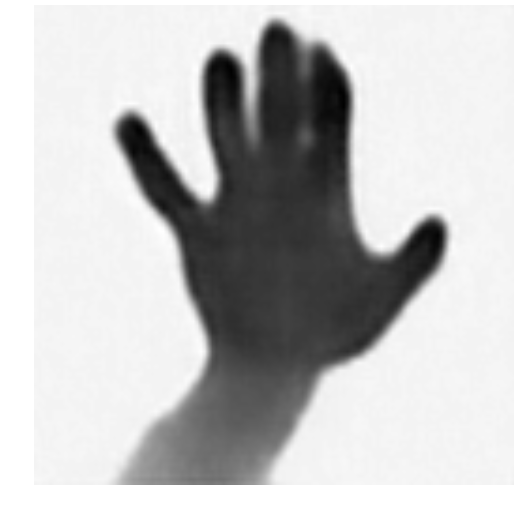} &
\includegraphics[width=0.19\linewidth]{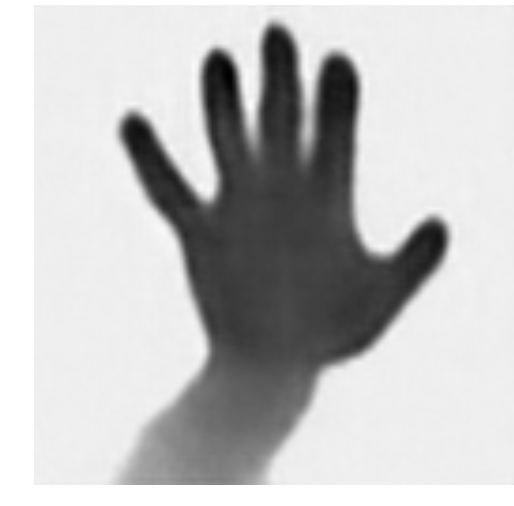} &
\includegraphics[width=0.17\linewidth]{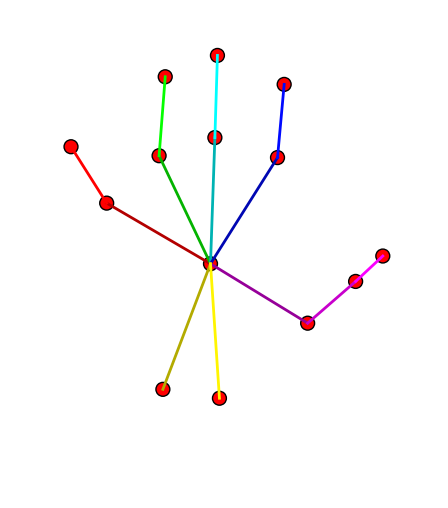} 
\end{tabular}
\end{center}
 \caption{Comparison with image-based pose optimization. \textbf{(Top)} results for image-based optimization, and \textbf{(bottom)} for our proposed method. From left to right: input depth image with initial pose, synthesized image for initial pose, after first, second iteration, and final pose. Minimizing the difference between the synthesized and the input image does not induce better poses. Thanks to the updater, our method can fit a good estimate. (Best viewed on screen)}
\label{fig:gensamplesBFGS}
\end{figure}
\egroup

\subsection{Qualitative results}

\bgroup
\setlength{\tabcolsep}{1pt}
\begin{figure*}[t]
\begin{center}
\newlength{\sampimgw}
\setlength\sampimgw{0.12\linewidth}
\begin{tabular}{@{}cc|cc|cc|cc@{}}
\includegraphics[width=\sampimgw,trim={5cm 3cm 5cm 1cm},clip]{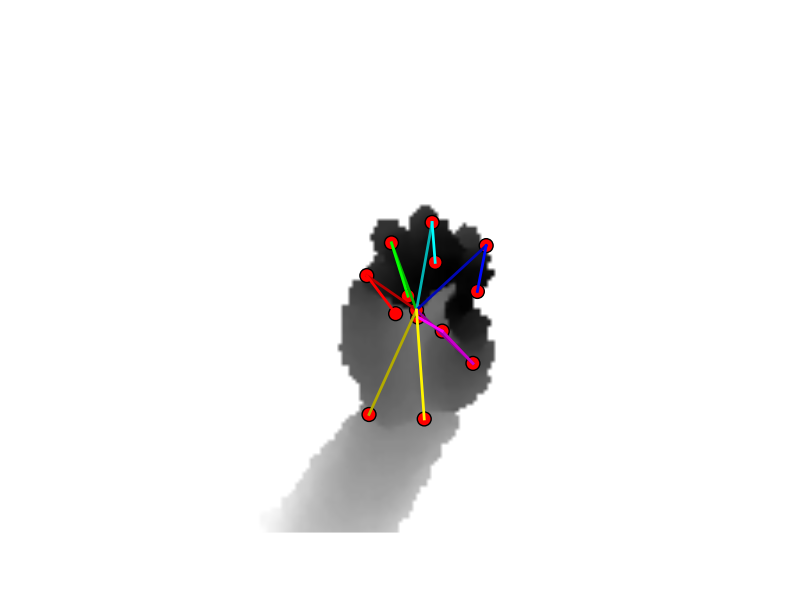} &
\includegraphics[width=\sampimgw,trim={5cm 3cm 5cm 1cm},clip]{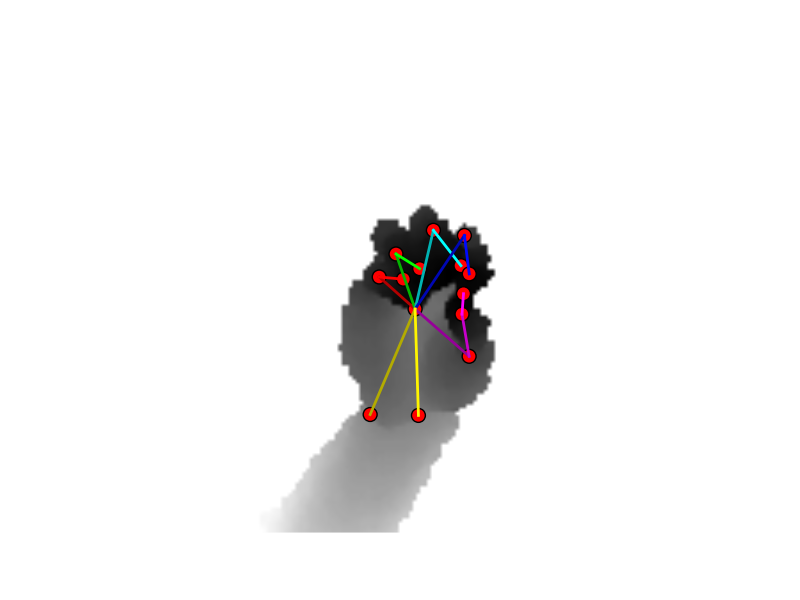} &
\includegraphics[width=\sampimgw,trim={5cm 3cm 5cm 1cm},clip]{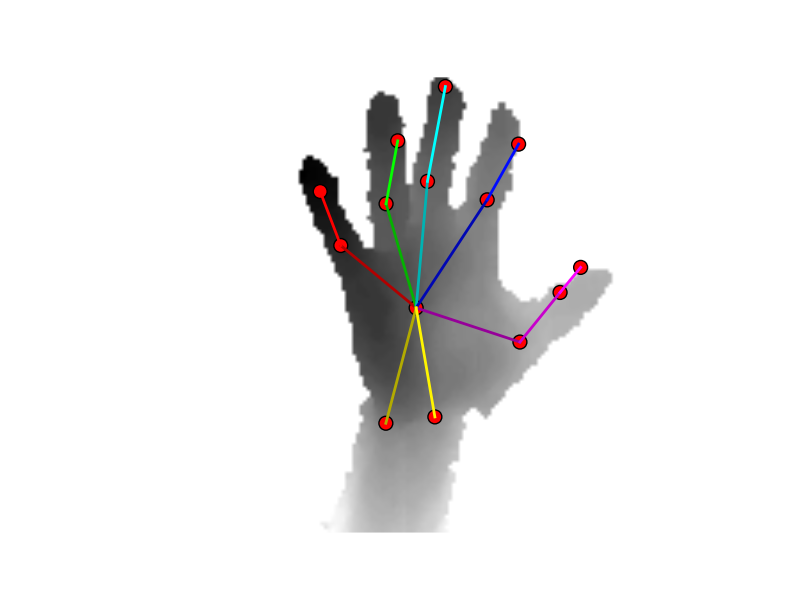} &
\includegraphics[width=\sampimgw,trim={5cm 3cm 5cm 1cm},clip]{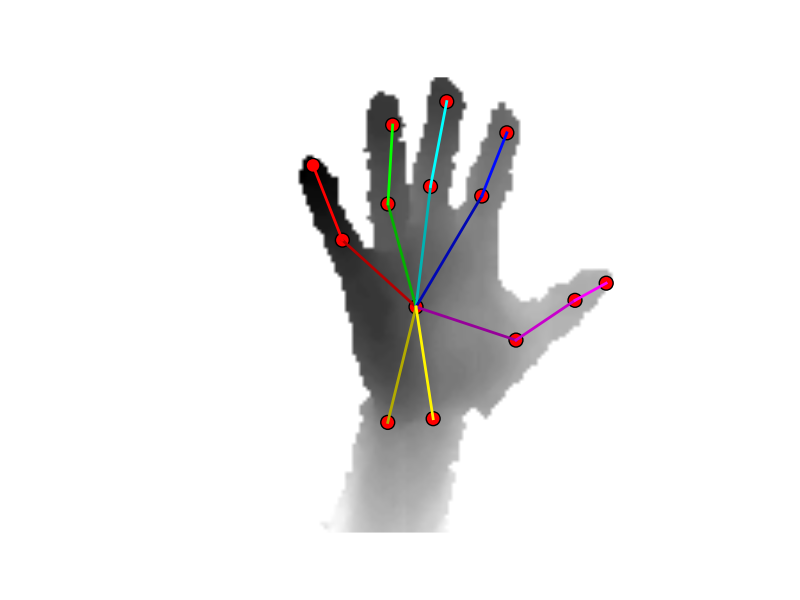} &
\includegraphics[width=\sampimgw,trim={5cm 3cm 5cm 1cm},clip]{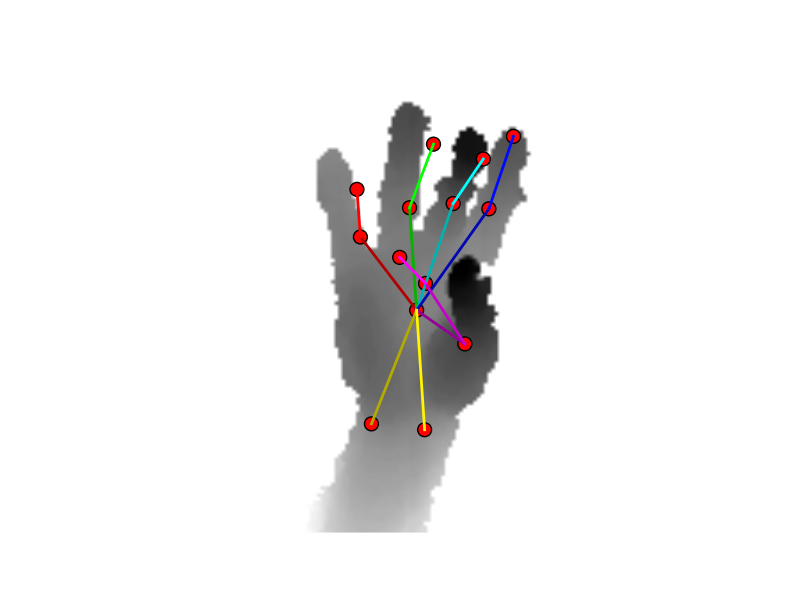} &
\includegraphics[width=\sampimgw,trim={5cm 3cm 5cm 1cm},clip]{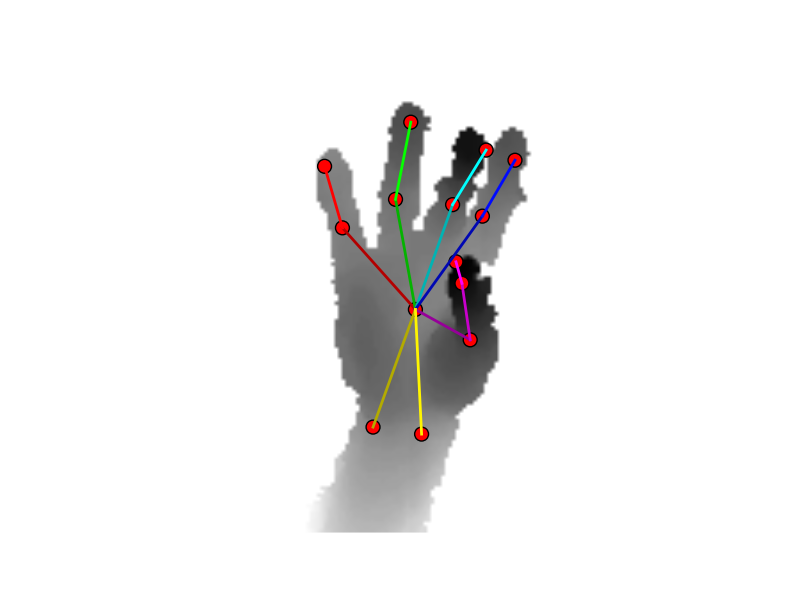} &
\includegraphics[width=\sampimgw,trim={5cm 1cm 3cm 1cm},clip]{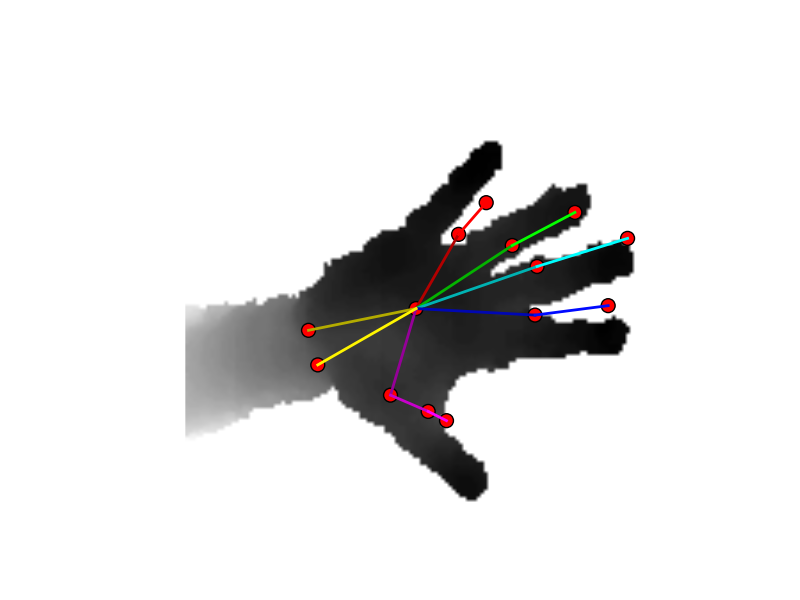} &
\includegraphics[width=\sampimgw,trim={5cm 1cm 3cm 1cm},clip]{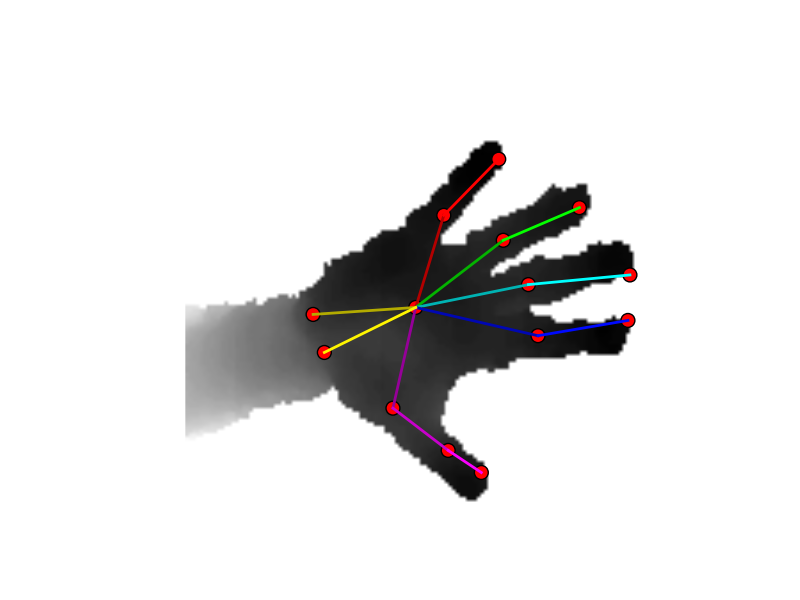} \\

\midrule
\includegraphics[width=\sampimgw,trim={5cm 1cm 5cm 5cm},clip]{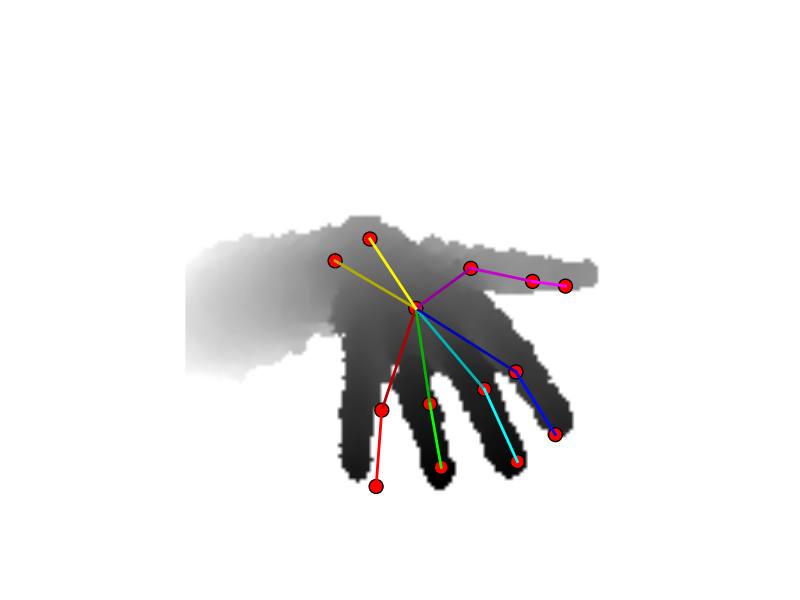} &
\includegraphics[width=\sampimgw,trim={5cm 1cm 5cm 5cm},clip]{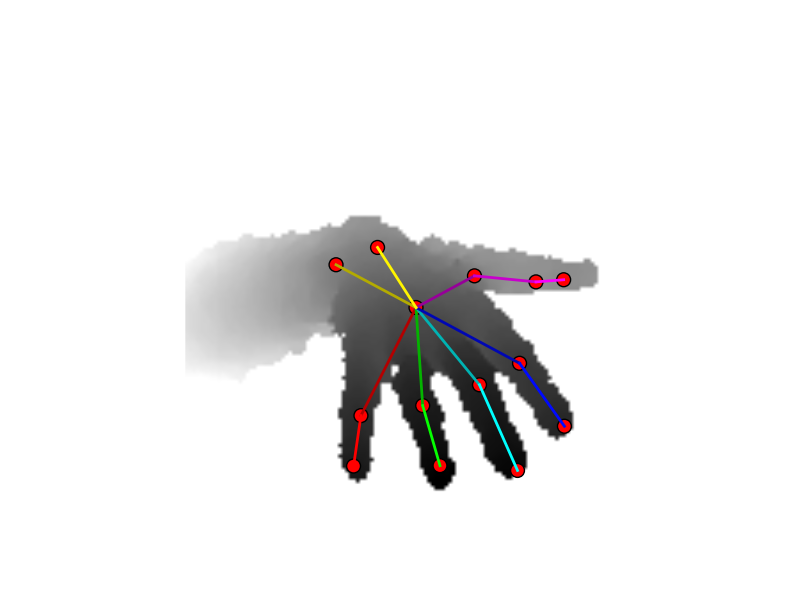} &
\includegraphics[width=\sampimgw,trim={5cm 3cm 5cm 1cm},clip]{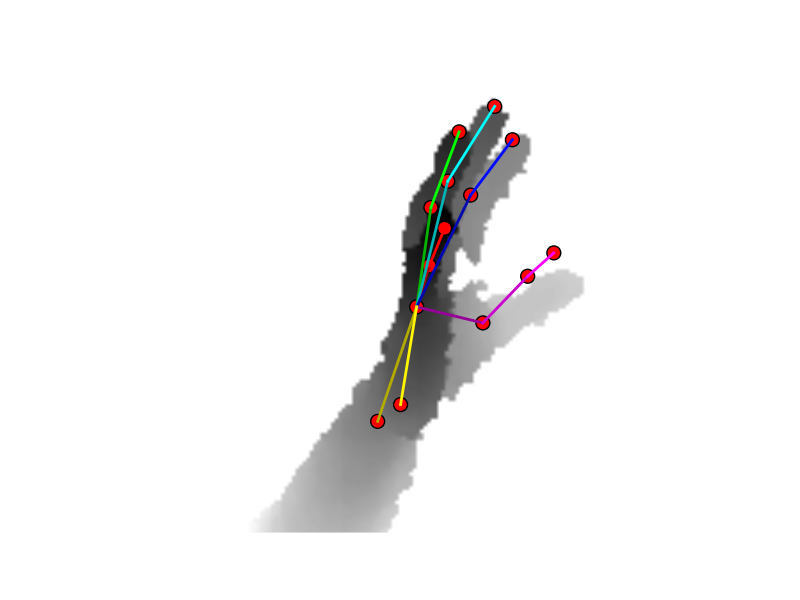} &
\includegraphics[width=\sampimgw,trim={5cm 3cm 5cm 1cm},clip]{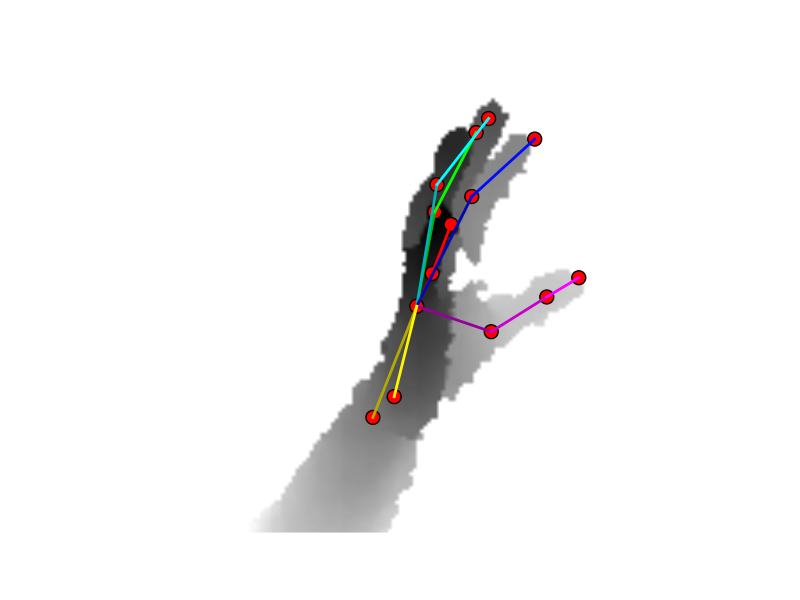} &
\includegraphics[width=\sampimgw,trim={5cm 3cm 5cm 1cm},clip]{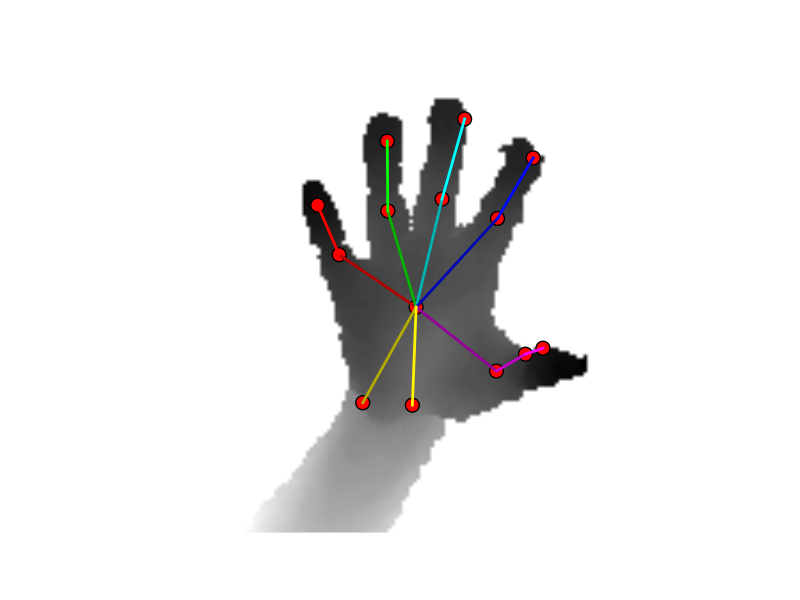} &
\includegraphics[width=\sampimgw,trim={5cm 3cm 5cm 1cm},clip]{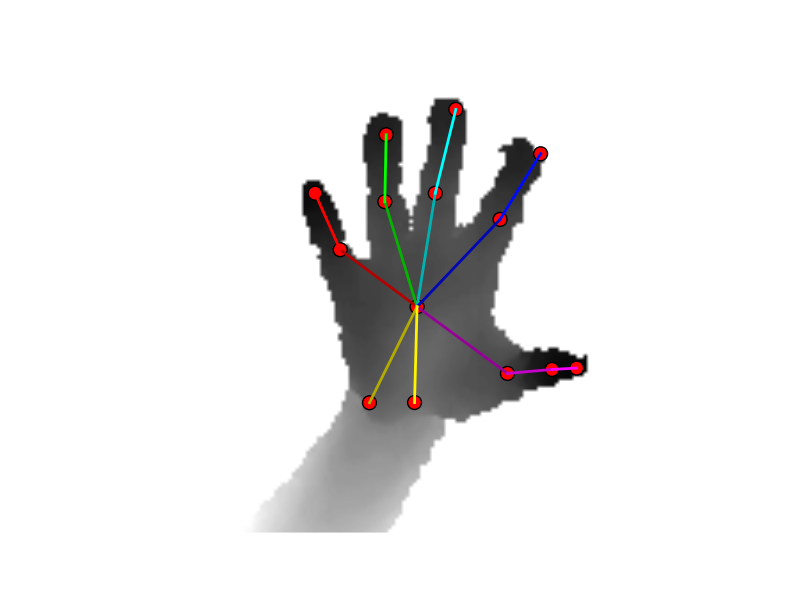} &
\includegraphics[width=\sampimgw,trim={5cm 3cm 5cm 1cm},clip]{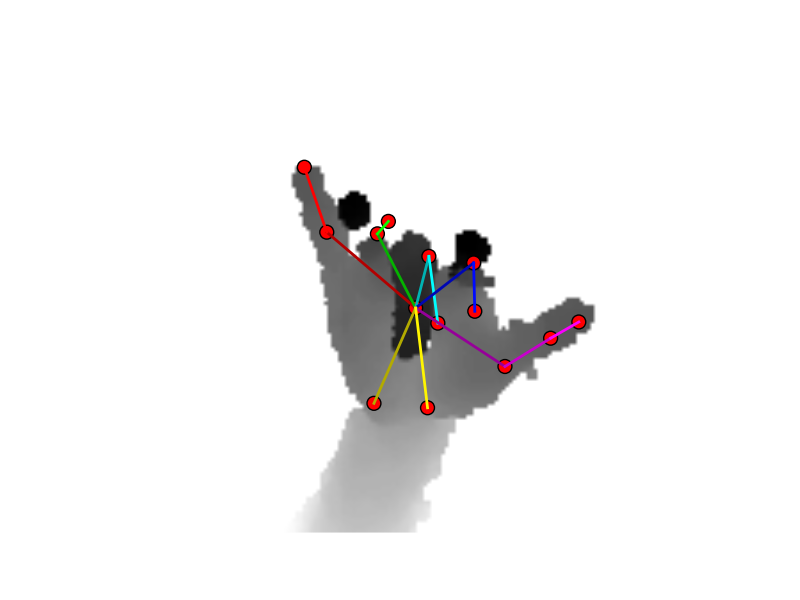} &
\includegraphics[width=\sampimgw,trim={5cm 3cm 5cm 1cm},clip]{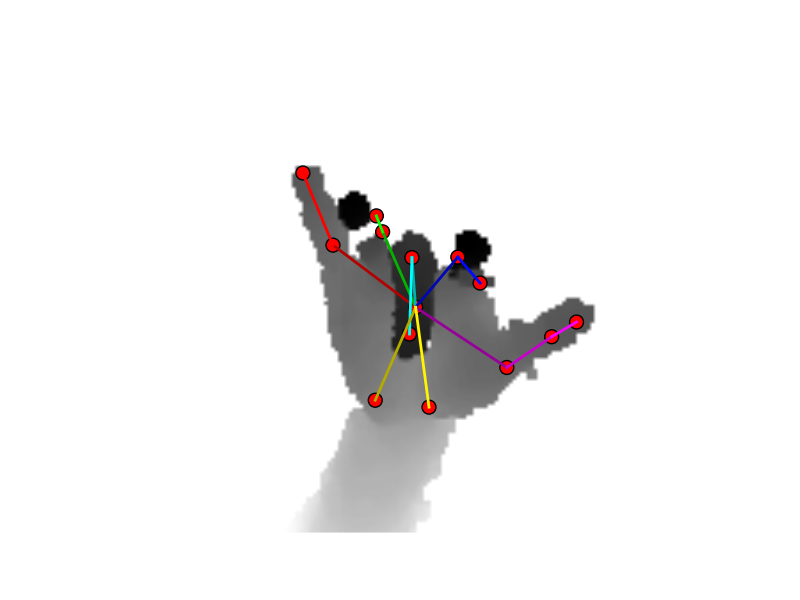} 

\end{tabular}
\end{center}
 \caption{Qualitative results for NYU dataset. We show the inferred joint locations on the depth images where the depth is color coded in gray-scale. The individual fingers are color coded, where the bones of each finger share the same color, but with a different hue. The left image of each pair shows the initialization and the right image shows the pose after applying our method. Our method applies to a wide variety of poses and is tolerant to noise, occlusions and missing depth values as shown in several images. (Best viewed on screen)}
\label{fig:results_qualitative}
\vspace{-1.2em}
\end{figure*}
\egroup

Fig.~\ref{fig:results_qualitative} shows some qualitative examples. For some
examples, the predictor provides already a good pose, which we can still
improve, especially for the thumb. For worse initializations, also larger updates on the pose can be achieved by our proposed method, to better explain the evidence in the image.

\subsection{Runtime}

Our method is implemented in Python using the Theano library~\cite{Bergstra2010}
and we run the experiments on a computer equipped with an Intel Core i7, 16GB of
RAM, and an nVidia GeForce GTX 780 Ti GPU. Training takes about ten hours for
each CNN. 

The runtime is composed of the discriminative initialization that takes 0.07~ms,
the updater network takes 1.2~ms for each iteration, and that already
includes the synthesizer with 0.8~ms. In practice we iterate our updater twice, thus our method performs very fast at over 400~fps on a single GPU. The
runtime of our method compares favorably with other model-based methods ranging between 12 and
60~fps~\cite{Melax2013,Oikonomidis2011a,Qian2014,Sridhar2013,Xu2013}.

\section{Discussion and Conclusion}

While our approach is not really biologically-inspired, it should be noted that
similar feedback mechanisms also have the support of strong biological evidence. It has been
shown that feedback paths in the brain and especially in the visual cortex
actually consist of \emph{more} neurons than the forward
path~\cite{Dayan05}. Their functional role remains mostly
unexplained~\cite{Briggs08}; our approach could be a possible explanation in the
case of feedback in the visual cortex, but of course, this would need to be
proved.

It should also be noted that our predictor and our synthesizer are trained with
exactly the same data. One may then ask how our approach can improve the first
estimate made by the predictor. The combination of the synthesizer
and the updater network provides us with the possibility for simply yet considerably
augmenting the training data to learn the update of the pose: For a given input image,
we can draw arbitrary numbers of samples of poses through which the updater is then
trained to move closer to the ground-truth. In this way, we can explore regions
of the pose space which are not present in the training data, but might be
returned by the predictor when applied to unseen images.

Finally, our approach has nothing really specific to the hand pose detection or
the use of a depth camera, since all its components are learned, from the
prediction of the initialization to the generation of images and the update
computation. The representation of the pose itself is also very simple and does
not have to take into account the specifics of the structure of the hand. We
therefore believe that, given proper training data, our approach can be applied
to many detection and tracking problems and also with different sensors.

\paragraph*{Acknowledgements:} 
This work was funded by the Christian Doppler Laboratory for Handheld Augmented Reality and the TU Graz FutureLabs fund.

{\small
\bibliographystyle{ieee}
\bibliography{short,cleaned_biblio}

\begin{thebibliography}{10}\itemsep=-1pt

\bibitem{Ballan2012}
L.~Ballan, A.~Taneja, J.~Gall, L.~V. Gool, and M.~Pollefeys.
\newblock {Motion Capture of Hands in Action Using Discriminative Salient
  Points}.
\newblock In {\em ECCV}, 2012.

\bibitem{Bergstra2010}
J.~Bergstra, O.~Breuleux, F.~Bastien, P.~Lamblin, R.~Pascanu, G.~Desjardins,
  J.~Turian, D.~Warde-Farley, and Y.~Bengio.
\newblock {Theano: A {CPU} and {GPU} Math Expression Compiler}.
\newblock In {\em Proc. of SciPy}, 2010.

\bibitem{Bishop06}
C.~Bishop.
\newblock {\em {Pattern Recognition and Machine Learning}}.
\newblock Springer, 2006.

\bibitem{Briggs08}
F.~Briggs and W.~Usrey.
\newblock {Emerging Views of Corticothalamic Function}.
\newblock {\em Current Opinion in Neurobiology}, 2008.

\bibitem{Byrd1995}
R.~H. Byrd, P.~Lu, J.~Nocedal, and C.~Zhu.
\newblock {A Limited Memory Algorithm for Bound Constrained Optimization}.
\newblock {\em SIAM Journal on Scientific and Statistical Computing}, 16(5),
  1995.

\bibitem{Chopra2005}
S.~Chopra, R.~Hadsell, and Y.~LeCun.
\newblock {Learning a Similarity Metric Discriminatively, with Application to
  Face Verification}.
\newblock In {\em CVPR}, 2005.

\bibitem{Dayan05}
P.~Dayan and L.~Abbott.
\newblock {\em {Theoretical Neuroscience: Computational and Mathematical
  Modeling of Neural Systems}}.
\newblock {MIT Press}, 2005.

\bibitem{LaGorce2011}
M.~de~La~Gorce, D.~J. Fleet, and N.~Paragios.
\newblock {Model-Based 3D Hand Pose Estimation from Monocular Video}.
\newblock {\em PAMI}, 33(9), 2011.

\bibitem{Dosovitskiy2015}
A.~Dosovitskiy, J.~T. Springenberg, and T.~Brox.
\newblock {Learning to Generate Chairs with Convolutional Neural Networks}.
\newblock In {\em CVPR}, 2015.

\bibitem{Erol2007}
A.~Erol, G.~Bebis, M.~Nicolescu, R.~D. Boyle, and X.~Twombly.
\newblock {Vision-Based Hand Pose Estimation: A Review}.
\newblock {\em CVIU}, 108(1-2), 2007.

\bibitem{Goodfellow2014}
I.~J. Goodfellow, J.~Pouget-Abadie, M.~Mirza, B.~Xu, D.~Warde-Farley, S.~Ozair,
  A.~Courville, and Y.~Bengio.
\newblock {Generative Adversarial Nets}.
\newblock In {\em NIPS}, 2014.

\bibitem{Jain2014}
A.~Jain, J.~Tompson, M.~Andriluka, G.~W. Taylor, and C.~Bregler.
\newblock Learning human pose estimation features with convolutional networks.
\newblock In {\em Proc. of ICLR}, 2014.

\bibitem{Keskin2011}
C.~Keskin, F.~K{\i}ra{\c{c}}, Y.~E. Kara, and L.~Akarun.
\newblock {Real Time Hand Pose Estimation Using Depth Sensors}.
\newblock In {\em ICCV}, 2011.

\bibitem{Keskin2012}
C.~Keskin, F.~K{\i}ra{\c{c}}, Y.~E. Kara, and L.~Akarun.
\newblock {Hand Pose Estimation and Hand Shape Classification Using
  Multi-Layered Randomized Decision Forests}.
\newblock In {\em ECCV}, 2012.

\bibitem{Kulkarni2015}
T.~D. Kulkarni, W.~Whitney, P.~Kohli, and J.~B. Tenenbaum.
\newblock {Deep Convolutional Inverse Graphics Network}.
\newblock In {\em NIPS}, 2015.

\bibitem{Kulkarni2014}
T.~D. Kulkarni, I.~Yildirim, P.~Kohli, W.~A. Freiwald, and J.~B. Tenenbaum.
\newblock {Deep Generative Vision as Approximate Bayesian Computation}.
\newblock In {\em NIPS}, 2014.

\bibitem{Kuznetsova2013}
A.~Kuznetsova, L.~Leal-taixe, and B.~Rosenhahn.
\newblock {Real-Time Sign Language Recognition Using a Consumer Depth Camera}.
\newblock In {\em ICCV}, 2013.

\bibitem{Liu2015}
S.~Liu, X.~Liang, L.~Liu, X.~Shen, J.~Yang, C.~Xu, L.~Lin, X.~Cao, and S.~Yan.
\newblock {Matching-CNN Meets KNN: Quasi-Parametric Human Parsing}.
\newblock In {\em CVPR}, 2015.

\bibitem{Melax2013}
S.~Melax, L.~Keselman, and S.~Orsten.
\newblock {Dynamics Based 3D Skeletal Hand Tracking}.
\newblock In {\em Proc. of Graphics Interface Conference}, 2013.

\bibitem{Nair2008}
V.~Nair, J.~Susskind, and G.~E. Hinton.
\newblock {Analysis-By-Synthesis by Learning to Invert Generative Black Boxes}.
\newblock In {\em Proc. of ICANN}, 2008.

\bibitem{Oberweger2015}
M.~Oberweger, P.~Wohlhart, and V.~Lepetit.
\newblock {Hands Deep in Deep Learning for Hand Pose Estimation}.
\newblock In {\em Proc. of CVWW}, 2015.

\bibitem{Oikonomidis2011a}
I.~Oikonomidis, N.~Kyriazis, and A.~A. Argyros.
\newblock {Efficient Model-Based 3D Tracking of Hand Articulations Using
  Kinect}.
\newblock In {\em BMVC}, 2011.

\bibitem{Oikonomidis2011}
I.~Oikonomidis, N.~Kyriazis, and A.~A. Argyros.
\newblock {Full DOF Tracking of a Hand Interacting with an Object by Modeling
  Occlusions and Physical Constraints}.
\newblock In {\em ICCV}, 2011.

\bibitem{Plaenkers03}
R.~Pl{\"a}nkers and P.~Fua.
\newblock {Articulated Soft Objects for Multi-View Shape and Motion Capture}.
\newblock {\em PAMI}, 25(10), 2003.

\bibitem{Primesense2015}
PrimeSense.
\newblock {Primesense 3D Sensors}, 2015.

\bibitem{Qian2014}
C.~Qian, X.~Sun, Y.~Wei, X.~Tang, and J.~Sun.
\newblock {Realtime and Robust Hand Tracking from Depth}.
\newblock In {\em CVPR}, 2014.

\bibitem{Scherer2010}
D.~Scherer, A.~M{\"u}ller, and S.~Behnke.
\newblock {Evaluation of Pooling Operations in Convolutional Architectures for
  Object Recognition}.
\newblock In {\em Proc. of ICANN}, 2010.

\bibitem{Sharp2015}
T.~Sharp, C.~Keskin, D.~Robertson, J.~Taylor, J.~Shotton, D.~Kim, C.~Rhemann,
  I.~Leichter, A.~Vinnikov, Y.~Wei, D.~Freedman, P.~Kohli, E.~Krupka,
  A.~Fitzgibbon, and S.~Izadi.
\newblock {Accurate, Robust, and Flexible Real-Time Hand Tracking}.
\newblock In {\em Proc. of CHI}, 2015.

\bibitem{Sridhar2013}
S.~Sridhar, A.~Oulasvirta, and C.~Theobalt.
\newblock {Interactive Markerless Articulated Hand Motion Tracking Using RGB
  and Depth Data}.
\newblock In {\em ICCV}, 2013.

\bibitem{Tang2014}
D.~Tang, H.~J. Chang, A.~Tejani, and T.-K. Kim.
\newblock {Latent Regression Forest: Structured Estimation of 3D Articulated
  Hand Posture}.
\newblock In {\em CVPR}, 2014.

\bibitem{Tang2013}
D.~Tang, T.~Yu, and T.~Kim.
\newblock {Real-Time Articulated Hand Pose Estimation Using Semi-Supervised
  Transductive Regression Forests}.
\newblock In {\em ICCV}, 2013.

\bibitem{Tang2014a}
Y.~Tang, N.~Srivastava, and R.~Salakhutdinov.
\newblock {Learning Generative Models with Visual Attention}.
\newblock In {\em NIPS}, 2014.

\bibitem{Taylor2012}
J.~Taylor, J.~Shotton, T.~Sharp, and A.~Fitzgibbon.
\newblock {The Vitruvian Manifold: Inferring Dense Correspondences for One-Shot
  Human Pose Estimation}.
\newblock In {\em CVPR}, 2012.

\bibitem{Tieleman2012}
T.~Tieleman and G.~Hinton.
\newblock {Lecture 6.5-Rmsprop: Divide the Gradient by a Running Average of Its
  Recent Magnitude}.
\newblock {\em COURSERA: Neural Networks for Machine Learning}, 2012.

\bibitem{Tompson2014}
J.~Tompson, M.~Stein, Y.~LeCun, and K.~Perlin.
\newblock {Real-Time Continuous Pose Recovery of Human Hands Using
  Convolutional Networks}.
\newblock {\em ACM Transactions on Graphics}, 33, 2014.

\bibitem{Tzionas2014}
D.~Tzionas, A.~Srikantha, P.~Aponte, and J.~Gall.
\newblock {Capturing Hand Motion with an RGB-D Sensor, Fusing a Generative
  Model with Salient Points}.
\newblock In {\em Proc. of GCPR}, 2014.

\bibitem{Xu2013}
C.~Xu and L.~Cheng.
\newblock {Efficient Hand Pose Estimation from a Single Depth Image}.
\newblock In {\em ICCV}, 2013.

\bibitem{Zeiler2014}
M.~D. Zeiler and R.~Fergus.
\newblock {Visualizing and Understanding Convolutional Networks}.
\newblock In {\em ECCV}, 2014.

\bibitem{Zeiler2011}
M.~D. Zeiler, G.~W. Taylor, and R.~Fergus.
\newblock {Adaptive Deconvolutional Networks for Mid and High Level Feature
  Learning}.
\newblock In {\em ICCV}, 2011.

\end{thebibliography}
}

\end{document}